\documentclass[11pt]{article}

\usepackage[margin=1in]{geometry}
\usepackage{amsmath,amssymb,amsfonts}
\usepackage{graphicx}
\usepackage{booktabs}
\usepackage{longtable}
\usepackage{array}
\usepackage{multirow}
\usepackage{caption}
\usepackage{subcaption}
\usepackage[numbers,sort&compress]{natbib}
\usepackage{hyperref}
\usepackage{xcolor}
\usepackage{enumitem}
\usepackage{microtype}
\usepackage{placeins}
\usepackage{float}

\hypersetup{
  pdftitle={Relational Rank Geometry in Transformers: Detecting and Steering Hidden-State Relation Frames},
  pdfauthor={Mazen Kobrosly},
  pdfkeywords={mechanistic interpretability, transformer representations, relation frames, Plucker coordinates, activation steering},
  colorlinks=true,
  linkcolor=blue,
  citecolor=blue,
  urlcolor=blue
}

\setlist[itemize]{leftmargin=1.25em}
\setlist[enumerate]{leftmargin=1.25em}
\newcommand{\R}{\mathbb{R}}
\newcommand{\Hpm}{H_{\pm}}

\newcommand{\clean}{\mathrm{clean}}
\newcommand{\corr}{\mathrm{corrupt}}
\newcommand{\patch}{\mathrm{patch}}

\newcommand{\sgn}{\operatorname{sign}}

\title{Relational Rank Geometry in Transformers:\\Detecting and Steering Hidden-State Relation Frames}
\author{Mazen Kobrosly\\Independent Researcher\\mazenkobrosly@gmail.com}
\date{}

\begin{document}
\maketitle

\begin{abstract}
Transformer hidden states are usually interpreted through local or low-order objects: a neuron, a sparse feature, an attention head, a residual-stream direction, or an activation patch. This paper studies a complementary object: the rank-indexed geometry of relations among token tuples. I use Pl\"ucker sign entropy to measure whether controlled $r$-argument relations leave arity-matched orientation signatures in transformer hidden-state space. Across Llama-family 8B, 70B, and 405B checkpoints, true relation tuples show stronger orientation-sign consistency at the expected rank $k=r$ for $r=3,\ldots,6$, relative to scrambled relation tuples under matched random-control audits. Multi-template audits show that the effects survive surface variation, with all tested 405B rows retaining a positive expected-rank margin, while 8B and 70B retain positive rows with constructor-specific mixed cells. I then ask whether the same relation geometry can be steered. To test this, I use an edge-grid clean/corrupt intervention assay over 32 prompts, where the row/column scaffold and answer format stay fixed, the YES/NO relation map changes, and the corrupt hidden-state relation frame is patched toward clean or placebo targets. In the 70B and 405B runs, clean-targeted relation-frame paths recover clean-answer behavior and residual relation geometry, while centroid-only and equal-norm controls show negligible recovery. Site/order controls in 70B and 405B further separate marker-site importance from ordered clean-frame geometry, as target clean shape and cross-prompt clean shape recover behavior and residual geometry at the marker interface, whereas corrupt-donor transfer, same-site permutation/reflection, wrong-site clean deltas, centroid-only motion, and equal-norm noise fail or remain far below the clean-frame paths. The result is a controlled bridge from relation probing to relation-frame intervention: relation rank geometry can be detected, targeted, and behaviorally validated in transformer hidden states.
\end{abstract}

\section{Introduction}

Many tasks given to transformers depend not only on the individual tokens in a given prompt, but on how several tokens fit together. Yet the field of mechanistic interpretability is largely organized around local or low-order objects: individual neurons, attention heads, sparse features, residual-stream directions, circuit subgraphs, or activation patches \citep{elhage2021framework,bricken2023monosemanticity,cunningham2024sparse,wang2023interpretability,conmy2023acdc}. These tools are essential for studying localized features, components, and patches, but they leave open a basic question: when a model represents a relation among several tokens, what is the geometry of that relation as a whole?

Recent work makes this question sharper. A growing representation-geometry literature argues that many model variables are not well described as single directions. Rather, concepts can form curved manifolds, sparse autoencoders may fragment manifold structure across features, and geometry-aware steering can produce more faithful behavioral trajectories than straight-line activation steering \citep{bhalla2026saemanifolds,wurgaft2026manifoldsteering,sarfati2026beliefs,tiblias2026smds}. In parallel, relational-binding work on language models shows that they can encode entity--relation structure in organized subspaces or grid-like cell representations, while relational-complexity evaluations show that higher-arity binding remains a systematic difficulty for frontier models \citep{dai2026cellbinding,fesser2026rel}. This leaves room for an object between isolated features and unconstrained activation clouds, one that treats the relation itself as the unit of analysis.

This paper studies one such object, the relation frame, defined as the hidden-state configuration of token tuples that participate in a relation. I ask whether these tuples have a rank-indexed orientation signature, and whether the corresponding hidden-state frame can be targeted by intervention. The geometric machinery comes from Pl\"ucker coordinates, a classical way of representing subspaces by their determinant minors, widely used in projective geometry, Grassmannian geometry, and oriented matroid theory. I borrow this mathematical language as it gives a principled way to describe the orientation structure of several vectors at once, including how the full configuration changes when the participating tokens are reordered or scrambled. To translate this into a measurable statistic for hidden states, I introduce Pl\"ucker sign entropy, a diagnostic built from the signs of Pl\"ucker-style determinant minors. After projecting hidden states into a fixed low-dimensional analysis subspace, I then compute determinant signs of selected token-tuple minors and ask whether true relation tuples show lower sign entropy than scrambled controls under a matched random-control construction.

In the controlled arity prompts, the number of arguments bound by the relation is fixed by construction. This experimental design gives the diagnostic a direct rank target. If an $r$-argument relation is represented as an ordered multi-token configuration, its strongest orientation signature should appear at Pl\"ucker rank $k=r$. This does not require lower ranks to vanish, however, as higher-order relations can also produce measurable lower-$k$ traces that coexist with the expected $k=r$ enrichment. The test is whether the expected rank $k=r$ is enriched relative to the other measured ranks within the full rank profile.

Across Llama-family 8B, 70B, and 405B checkpoints, the expected Pl\"ucker rank is consistently enriched for controlled $r$-argument relations with $r=3,4,5,6$. The pattern is most uniform in 405B, where multi-template audits preserve positive diagonal margins across both argument-only and predicate-plus-argument constructors. Smaller checkpoints retain broad positive support, but with more template- and constructor-specific variation. The diagnostic result is therefore not a claim of uniform rank structure everywhere; it is evidence that relation rank is measurable and systematically modulated by model scale and prompt construction.

The diagnostic and intervention halves of the paper use the same class of underlying object, a relation frame, at different levels of analysis. Pl\"ucker sign entropy asks whether relation-bound token tuples have unusually consistent orientation structure. Steering treats the corresponding hidden-state configuration as a relation frame whose centroid, relative shape, and spanned subspace can be moved or controlled. The diagnostic motivates treating the tuple as a geometric object, while the steering experiments ask whether moving that object changes behavior.

To make this intervention test clean, I create an edge-grid assay that compares clean and corrupt relation states. Clean and corrupt prompts share the same row/column scaffold and answer format, but differ in their YES/NO relation markers. In 70B and 405B, clean prompts reliably elicit the clean answer, and corrupt prompts reliably elicit the corrupt answer. This gives a controlled setting in which the hidden-state relation frame can be moved from corrupt toward clean while answer behavior and residual geometry are measured along the path.

In the steering experiments, relation-frame shape carries the main effect. At patch layer 5, moving the corrupt relation-marker cloud toward the clean relation-marker cloud recovers both clean-answer behavior and residual relation geometry at readout layer 35. Shape-only, centroid-plus-shape, and Grassmann shape paths produce strong recovery, while centroid-only and equal-norm controls do not. A site-and-ordering audit then tests whether this success is simply a consequence of patching important YES/NO marker positions. In both 70B and 405B, target clean shape and cross-prompt clean shape recover behavior and residual geometry at the same marker sites, while corrupt-donor shapes and broken-frame or wrong-site placebos remain far below.

The steering results do not claim to identify a complete circuit. They test a simpler claim: whether a relation-level hidden-state object can be used as an intervention target. That distinction matters because an internal representation can be predictive without yielding reliable output correction when intervened on \citep{basu2026actionability}. The edge-grid assay therefore asks whether relation-frame geometry is merely visible in hidden states or whether clean-aligned changes to that geometry can recover the model's behavior.

Within this context, the experiments make three concrete contributions:
\begin{itemize}
\item a rank-indexed diagnostic assay showing that controlled $r$-argument relation tuples have stronger orientation-sign consistency at the expected Pl\"ucker rank $k=r$ across Llama-family 8B, 70B, and 405B checkpoints;
\item a competence-gated edge-grid assay comparing clean and corrupt states, which isolates relation-marker geometry from the surrounding row/column scaffold;
\item an interventional result showing that relation-frame shape, not centroid motion or matched-norm noise, recovers behavior and residual geometry in 70B and 405B, with 70B/405B site-and-ordering controls separating marker-site importance from clean-state ordered geometry.
\end{itemize}

The core claim is that relational rank geometry is both detectable by determinant-sign statistics and, through controlled relation tasks, targetable by hidden-state interventions that recover relation-dependent behavior.

\begin{figure}[!t]
\centering
\makebox[\linewidth][c]{\includegraphics[width=1.13\linewidth]{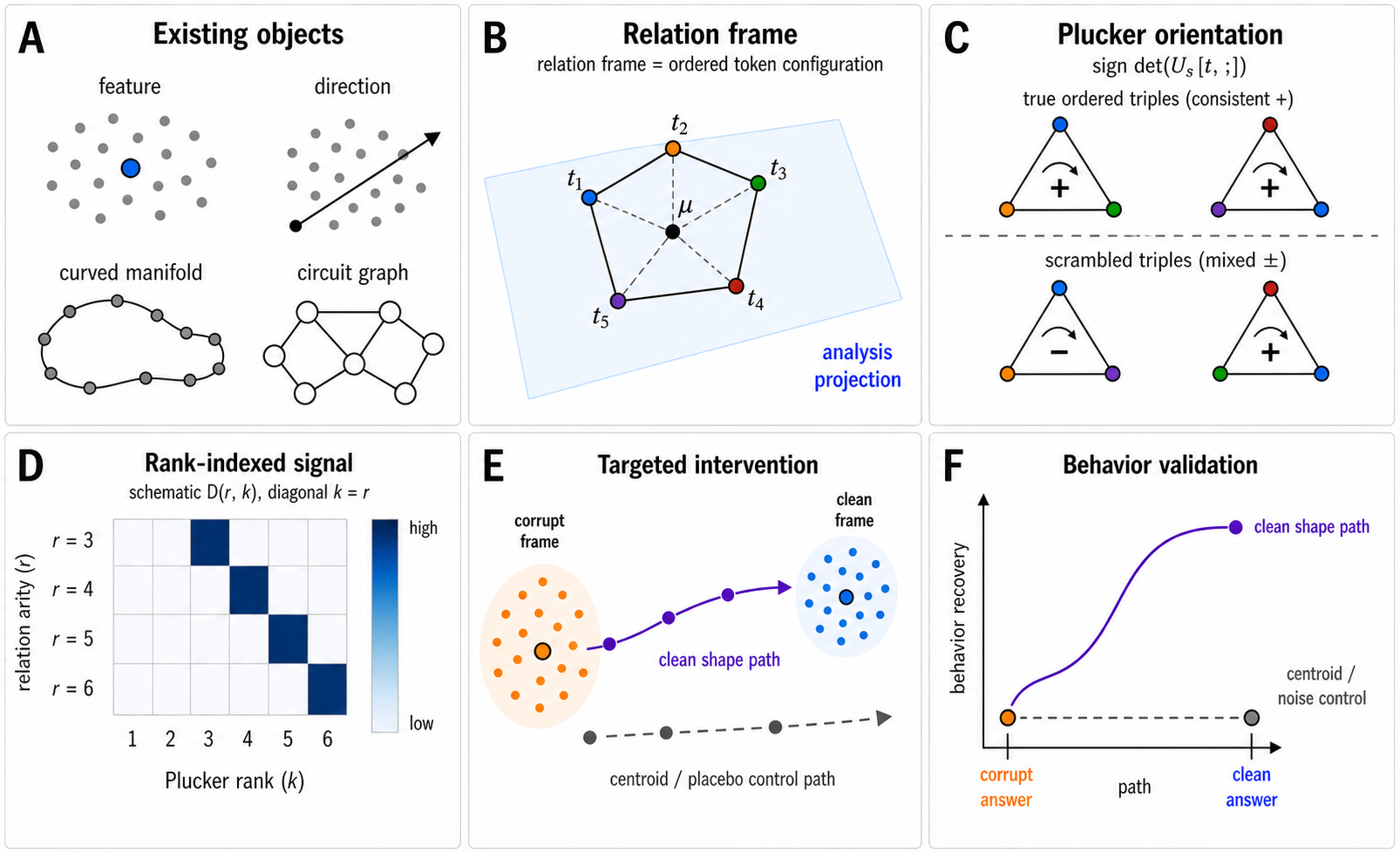}}
\vspace{-0.55em}
\caption{Relation frames as geometric intervention objects. The figure contrasts existing interpretability objects such as local features, directions, manifolds, and circuit graphs with the relation frame studied here. A relation frame is a finite ordered configuration of token hidden states participating in a relation, with a centroid, relative shape, and projected orientation structure. Pl\"ucker sign entropy measures orientation consistency over selected token tuples, producing a rank-indexed diagnostic over relation arity and Pl\"ucker rank. The steering panels then show the same class of object as a corrupt-to-clean intervention target, where clean shape paths recover behavior while centroid or placebo controls remain low.}
\label{fig:concept}
\end{figure}

\section{Pl\"ucker Sign Entropy and Relation Frames}

Pl\"ucker sign entropy tests whether selected relation tuples form configurations with consistent orientation signs after the hidden states are projected into a fixed low-dimensional analysis subspace.

For a $k$-token tuple, the relevant object is the sign of a row-selected determinant minor after projecting the hidden states and taking a rank-$k$ left-singular token embedding. Geometrically, this determinant sign records the orientation (or handedness) of the selected tuple in that analysis coordinate system while discarding scale. For $k=2$, this is the signed area of an ordered pair in the rank-2 token embedding; for $k=3$, it is the signed volume of an ordered triple in the rank-3 token embedding; higher $k$ uses the same ordered-minor sign structure. If true relation tuples repeatedly share one sign pattern while matched controls do not, the tuple set has lower sign entropy. The diagnostic therefore asks whether relation-bound tokens arrange into a consistent oriented configuration in the projected hidden-state coordinates.

Arity then gives the prediction a direct reading. If an $r$-argument relation is being represented as an $r$-way configuration, the strongest enrichment should appear at Pl\"ucker rank $k=r$ for argument-only tuples. This is the diagonal hypothesis tested below.

\subsection{Hidden-state coordinates}

To formalize this orientation structure, I first project the hidden states into a fixed low-dimensional analysis subspace. For a prompt and transformer layer $\ell$, let
\begin{equation}
H^{(\ell)} \in \R^{n \times d}
\end{equation}
be the matrix of token hidden states, where $n$ is the sequence length and $d$ is the model hidden dimension. The projected coordinates are
\begin{equation}
X = H^{(\ell)}P, \qquad P \in \R^{d \times p}.
\end{equation}
The main diagnostic analyses use a fixed Gaussian projection with $p=64$, with the projection seed held constant across reported comparisons. The projection matrix is sampled with i.i.d. Gaussian entries and scaled by $1/\sqrt{p}$; it is generated once per model dimension with the reported seed and then held fixed for matched comparisons. Projection-dimension and projection-seed audits are included in the replication bundle and are used as robustness checks against dependence on a single projection draw; they are not tabulated in the paper. All main-table comparisons hold the projection dimension and seed fixed. The steering runs use the same projection dimension and seed for residual-geometry readouts.

\begingroup
\setlength{\abovedisplayskip}{0.75\baselineskip}
\setlength{\belowdisplayskip}{0.70\baselineskip}
\setlength{\abovedisplayshortskip}{0.65\baselineskip}
\setlength{\belowdisplayshortskip}{0.70\baselineskip}

Within this projected space, compute
\begin{equation}
X = U\Sigma V^\top.
\end{equation}
For rank $k$, let $U_k \in \R^{n \times k}$ be the first $k$ left singular vectors. Given a token tuple $t=(i_1,\ldots,i_k)$, define
\begin{equation}
M_t = U_k[t,:] \in \R^{k \times k}, \qquad m_t = \det(M_t), \qquad s_t=\sgn(m_t).
\end{equation}
Operationally, the determinant is taken on the selected row minor of the top-$k$ left-singular token embedding $U_k$. It records the orientation of the selected token tuple after projecting into the prompt's top-$k$ singular coordinate system, with singular-value scale discarded. Because $U_k$ has orthonormal columns, this minor depends on the $k$-dimensional left singular subspace and the token-row selection rather than on raw hidden-state vector volume. Before computing determinants, each tuple is ordered by its fixed role positions in the prompt, making the sign statistic intentionally order-sensitive rather than permutation-invariant. The entropy is then computed from the signs of finite, nonzero determinants, with exactly zero minors dropped under the same rule for true, scrambled, and random tuple sets. Near-zero but nonzero minors are not thresholded in the headline statistic; the same rule is applied to all selector sets in each matched comparison. The sign entropy is invariant to global SVD column sign flips, since such flips multiply all selected determinants at a fixed prompt and rank by the same sign product and therefore only exchange the labels $+$ and $-$. This makes the readout an orientation-sign statistic rather than a volume statistic.

\begin{figure}[t]
\centering
\includegraphics[width=0.92\linewidth]{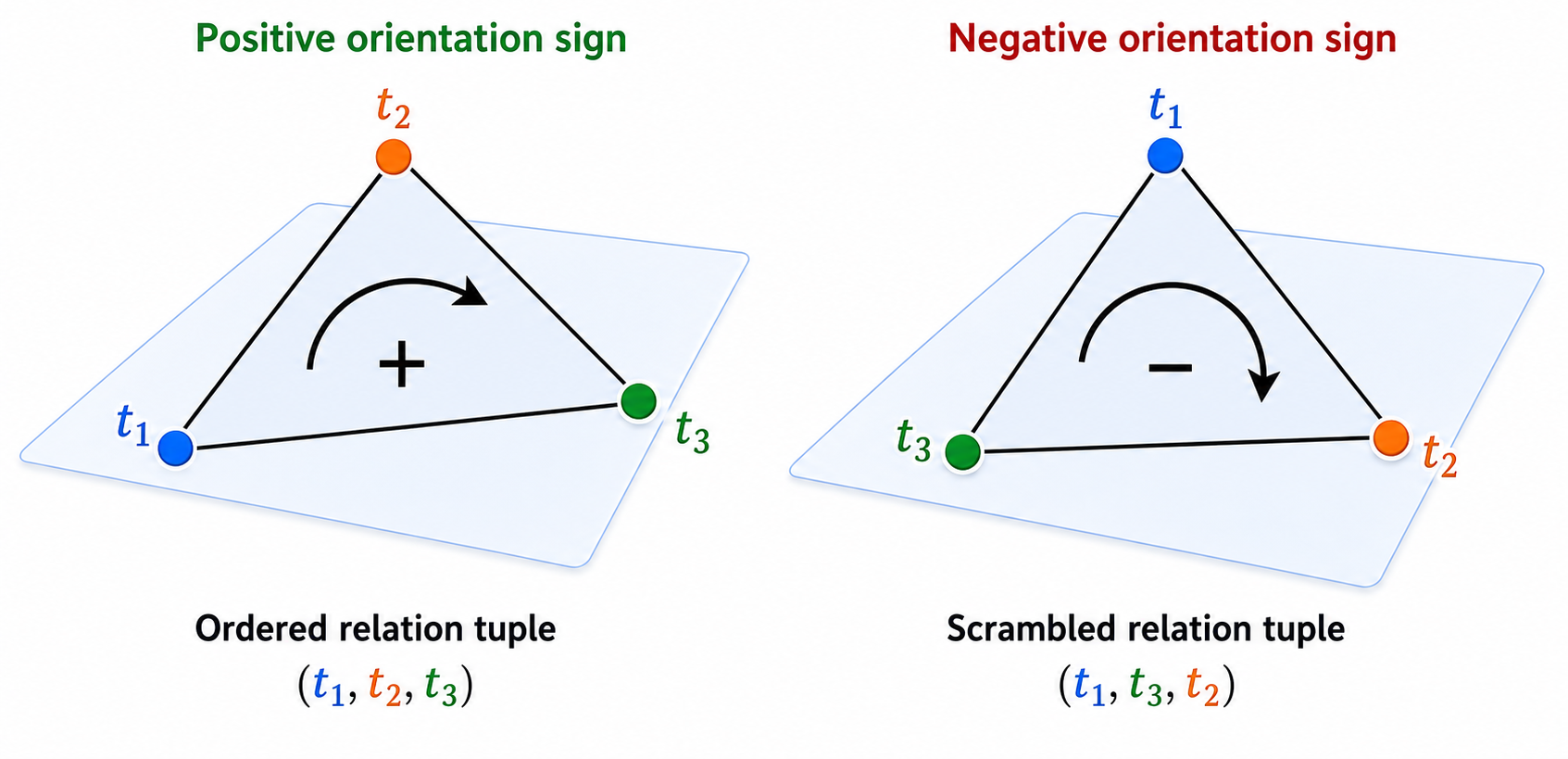}
\caption{Orientation-sign intuition for Pl\"ucker sign entropy in the $k=3$ case. After token hidden states are projected into an analysis subspace, an ordered triple has a determinant sign that records its orientation while discarding scale. Reordering the same three token vectors can flip this sign. Pl\"ucker sign entropy aggregates these signs over true, scrambled, and random tuple sets and asks whether relation-bound tuples have more consistent orientation signs than matched controls. Higher ranks use the same determinant-sign idea for larger projected token tuples.}
\label{fig:plucker-orientation-intuition}
\end{figure}

For a tuple set $T$, define sign entropy
\begin{equation}
\Hpm(T) = -\sum_{s \in \{-1,+1\}} p_s(T) \log_2 p_s(T),
\end{equation}
where $p_s(T)$ is the empirical frequency of sign $s$ in $T$. Lower sign entropy means the selected tuple set has more consistent orientation signs at rank $k$.

\subsection{Controlled arity gap and diagonal margin}

For controlled relational arity, define the random-minus-selector entropy gap
\begin{equation}
\Delta H_S(r,k)=\Hpm(T_{\mathrm{rand}}^{r,k})-\Hpm(T_S^{r,k}),
\end{equation}
where $S$ is either the true or scrambled selector. The controlled arity statistic is then
\begin{equation}
D(r,k)=\Delta H_{\mathrm{true}}(r,k)-\Delta H_{\mathrm{scrambled}}(r,k).
\end{equation}
A single matched random tuple set $T_{\mathrm{rand}}^{r,k}$ is used for the true and scrambled selectors within each prompt and rank. Thus the random term cancels algebraically in $D(r,k)$, giving $D(r,k)=\Hpm(T_{\mathrm{scrambled}}^{r,k})-\Hpm(T_{\mathrm{true}}^{r,k})$. I retain the random-minus-selector notation because the random baseline defines the matched-control construction and keeps the raw entropy gaps inspectable. Positive $D(r,k)$ means that true relation tuples have lower sign entropy than scrambled tuples at rank $k$.

The diagonal hypothesis is rank-specific and constructor-specific. Let $c$ denote the tuple constructor, with admissible rank set $\mathcal{K}_c(r)$ and expected rank $k_c^\star(r)$. For argument-only tuples, $\mathcal{K}_{\mathrm{arg}}(r)=\{1,\ldots,r\}$ and $k_{\mathrm{arg}}^\star(r)=r$. For predicate-plus-argument tuples, $\mathcal{K}_{\mathrm{pred+arg}}(r)=\{1,\ldots,r+1\}$ and $k_{\mathrm{pred+arg}}^\star(r)=r+1$. I use the same-layer constructor margin
\begin{equation}
\mathrm{margin}_c(r)=D_c\!\left(r,k_c^\star(r)\right)
-\max_{k\in\mathcal{K}_c(r),\, k \neq k_c^\star(r)}D_c(r,k),
\end{equation}
where $D_c$ is computed using the same constructor for true, scrambled, and random tuple sets. For the headline argument-only rows, this reduces to the simpler $D(r,r)-\max_{k\in\mathcal{K}_{\mathrm{arg}}(r),\,k\neq r}D(r,k)$ form. A positive diagonal margin means that the expected-rank cell is larger than the other admissible ranks at the selected layer. Notably, this does not require off-diagonal ranks to be inactive: higher-order relations can produce measurable lower-$k$ traces of pairwise or lower-order orientation structure that coexist with the expected-rank signal.

\subsection{Relation frames}

For steering, I use the term ``relation frame'' to mean the selected hidden-state configuration of the tokens that instantiate the relation in a prompt. In the edge-grid assay, the relation-marker cloud is the subset of this frame consisting of hidden states at selected YES/NO marker token positions. The visible rows, columns, and YES/NO markers are the prompt grid; the relation frame is the corresponding hidden-state object inside the model.

The frame language is useful because the same selected state has several controllable aspects: its centroid is the mean marker vector, its centered shape is the relative arrangement after subtracting that mean, and its subspace is the low-dimensional span of the centered shape. The steering paths used later move these aspects from corrupt toward clean in different ways. A linear path moves the marker vectors directly, a shape path replaces the relative arrangement while holding the corrupt centroid fixed, a Procrustes path rotates one centered cloud toward another, and a Grassmann path moves the subspace spanned by the cloud. These are operational handles on one finite hidden-state object, not separate claims about the full mechanism.

If $C_{\clean}$ and $C_{\corr}$ are the clean and corrupt relation-marker clouds at the patch layer, then a steering method defines a path
\begin{equation}
C(\alpha), \qquad \alpha \in [0,1],
\end{equation}
with $C(0)=C_{\corr}$. Clean-targeted paths set $C(1)$ to the clean relation-marker cloud or to a matched approximation of it. Control paths keep the same patching interface but replace the clean endpoint with centroid-only, noise, permutation, reflection, or random-subspace alternatives. The patch is applied to the corrupt prompt by replacing or shifting the hidden states at the selected YES/NO marker token positions according to the path. The readout then measures whether the answer and residual relation geometry move toward the clean state.

Pl\"ucker sign entropy reads order-sensitive orientation structure from selected token tuples after projection. Steering treats the corresponding selected hidden states as a patchable cloud. The two procedures therefore operate on the same kind of relation-frame object, an ordered role-indexed cloud of token hidden states selected by a relation, in different modes: first as a diagnostic object whose orientation signs are scored, and then as an intervention object whose shape is patched. They are not guaranteed to instantiate the same frame at the same layer; the bridge claim is that the same class of object is being measured and moved. The diagnostic asks whether the selected tuple has a consistent orientation signature; the steering assay asks whether moving the corresponding hidden-state configuration changes behavior.

Thus, the claim is intentionally state-level. A relation frame is not a complete decomposition of the upstream components that build it or the downstream components that consume it. The experiment's aim is focused on whether changing this selected relation state changes behavior and residual relation geometry.

\endgroup

\FloatBarrier

\section{Detecting Rank-Indexed Relation Structure}

\subsection{Controlled arity foundation}

Relational complexity gives a principled organizing axis for the prompt bank. In cognitive theory, arity distinguishes unary predicates, binary relations, ternary relations, and higher-order binding demands \citep{halford1998relational,gentner1983structure}. Recent LLM evaluations make this axis newly relevant: REL defines relational complexity as the number of independent entities or operands that must be simultaneously bound to apply a relation, and finds that model performance degrades as this complexity increases even when input size and entity count are controlled \citep{fesser2026rel}. Here, arity is not inferred from natural language after the fact. It is fixed by construction, since each controlled prompt is synthetically generated with a specified number of relation arguments. This section reports $r=3,4,5,6$, focusing on higher-order tuple binding under the same true, scrambled, and random comparisons.

In a ternary prompt, the true tuple is the set of three role tokens bound by a single relation instance. Scrambled tuples keep the same tokens available but recombine them so the binding is broken; random controls use matched tokens outside the true relation tuple. The question is whether the bound tuple has a more consistent orientation signature than these controls. I compute the statistic at the prompt level, so repeated tuple signs from the same prompt are not treated as independent evidence.

\begin{table}[!htb]
\centering
\small
\begin{tabular}{lcccccc}
\toprule
Model & $r$ & expected $k$ & layer & $D(r,r)$ & 95\% CI & margin \\
\midrule
8B & 3 & 3 & 25 & 0.640 & [0.617, 0.663] & 0.648 \\
8B & 4 & 4 & 20 & 0.594 & [0.577, 0.610] & 0.433 \\
8B & 5 & 5 & 20 & 0.826 & [0.813, 0.841] & 0.665 \\
8B & 6 & 6 & 20 & 0.953 & [0.940, 0.967] & 0.825 \\
70B & 3 & 3 & 40 & 0.306 & [0.289, 0.323] & 0.313 \\
70B & 4 & 4 & 55 & 0.473 & [0.458, 0.486] & 0.462 \\
70B & 5 & 5 & 40 & 0.644 & [0.623, 0.663] & 0.459 \\
70B & 6 & 6 & 40 & 0.397 & [0.367, 0.427] & 0.162 \\
405B & 3 & 3 & 30 & 0.114 & [0.098, 0.129] & 0.041 \\
405B & 4 & 4 & 30 & 0.228 & [0.206, 0.249] & 0.165 \\
405B & 5 & 5 & 30 & 0.632 & [0.594, 0.670] & 0.535 \\
405B & 6 & 6 & 40 & 0.375 & [0.352, 0.397] & 0.346 \\
\bottomrule
\end{tabular}
\caption{Controlled arity diagonal enrichment from the controlled arity bank. All rows report the expected-rank statistic $D(r,r)$ for argument-only tuple constructors, with prompt-level confidence intervals and same-layer diagonal margins. A margin can exceed the displayed $D(r,r)$ value when the largest off-diagonal admissible rank has negative $D$ at the selected layer. Multi-template generalization is reported separately in Table~\ref{tab:multitemplate}.}
\label{tab:arity-main}
\end{table}

Controlled arity provides the baseline diagnostic evidence. For each reported arity $r=3,\ldots,6$, true argument tuples have positive expected-rank enrichment $D(r,r)$ across the 8B, 70B, and 405B checkpoints. The same-layer diagonal margin is also positive in every row, indicating that the arity-matched cell is the largest tested rank at the selected layer. Table~\ref{tab:arity-main} reports the corresponding layers, confidence intervals, and margins. The displayed layer is a descriptive best-layer summary selected from within the audited layer grid, not a pre-registered layer. Appendix Table~\ref{tab:heldout} reports a deterministic held-out audit that selects layers on even prompt IDs and evaluates the diagonal rows on odd prompt IDs; held-out $D(r,r)$ remains positive with bootstrap intervals excluding zero for every reported model/arity row. The row-level pattern is heterogeneous: 8B is strong in this controlled bank, 70B is strongest at $r=5$ and remains positive for all reported arities, and 405B shows its largest reported margin at $r=5$.

\subsection{Multi-template generalization}

After the controlled bank establishes the diagonal test, the multi-template audit asks whether the signal is tied to relation structure rather than to one surface form of the prompt bank. A single template leaves open simpler explanations, such as the model responding to repeated wording, fixed role positions, or lexical cues that happen to correlate with arity. To test this, the audit uses arity-neutral predicates and several prompt templates for $r=3,\ldots,6$.

A second check varies the tuple constructor. The argument-only constructor selects the role tokens alone and therefore expects the diagonal at $k=r$. The predicate-plus-argument constructor prepends the predicate token, making the selected tuple one token larger and shifting the expected rank to $k=r+1$. This gives a sharper check: if the statistic is tracking the selected relation tuple, the expected rank should move with the constructor rather than staying fixed at a memorized arity label.

Combined, these variations extend the controlled-arity result beyond a single prompt family. The expected-rank signal persists under surface changes, separating the diagnostic from a template-specific artifact. The audit also reveals structured variation across models and tuple constructors. The 405B checkpoint gives the most consistent pattern, with positive diagonal margins in every tested argument-only and predicate-plus-argument row. The 70B audit is constructor-specific: argument-only rows have positive margins at $r=3,4,5$ and a near-zero positive margin at $r=6$, while predicate-plus-argument rows have a slightly negative margin at $r=3$ and positive margins at $r=4,5,6$. The diagnostic result is structured rank sensitivity: controlled relation tuples induce rank-indexed orientation structure whose strength varies with model scale, tuple constructor, and prompt surface.

\begin{figure}[!htb]
\centering
\makebox[\linewidth][c]{\includegraphics[width=1.12\linewidth]{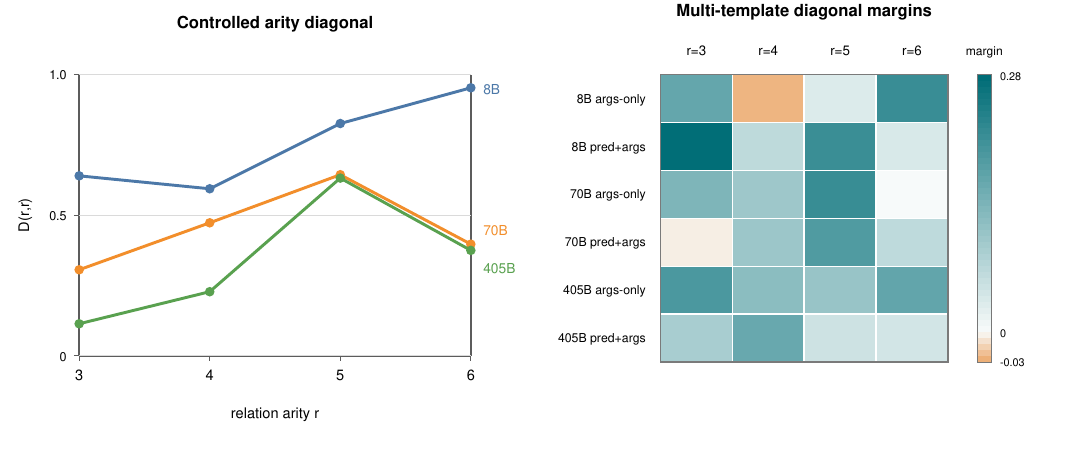}}
\caption{Diagnostic relation-rank evidence. Left: controlled-arity diagonal values $D(r,r)$ for relation arities $r=3,\ldots,6$, where $D(r,r)$ measures expected-rank orientation enrichment for argument-only tuples. Right: multi-template diagonal margins by checkpoint, arity, and tuple constructor. In the heatmap, args-only means the tuple contains only the relation arguments and expects $k=r$; pred+args means the predicate token is prepended to the arguments and expects $k=r+1$. Darker teal indicates larger positive expected-rank margins, while orange indicates negative cells. The 405B audit has positive margins across tested rows; 8B and 70B show constructor-specific mixed cells rather than uniform dominance.}
\label{fig:arity-multitemplate}
\end{figure}

\section{Edge-Grid Clean/Corrupt Assay}

Steering requires a task in which relation structure can change while most of the prompt remains fixed. I use an edge-grid assay built from $8 \times 8$ grids. Each prompt lists rows $A00$--$A07$ and columns $B00$--$B07$ in a fixed order. YES marks active edges; NO marks inactive edges. The relation state is the YES/NO pattern over this fixed row/column scaffold.

This construction yields clean and corrupt prompts that share the same vocabulary, answer format, and grid scaffold, but differ in their relation maps. In the clean run, the YES edges form a coherent map, for example the identity map in which each $A_i$ points to $B_i$, and the correct answer describes that clean map. In the corrupt run, the row/column scaffold is preserved but the YES/NO relation state is changed to a duplicate-target map in which multiple $A$ rows point to reused $B$ columns. The correct answer then changes with the relation state. The corrupt prompt is therefore a structured alternative relation state with its own matching answer.

The patched corrupt run starts from the corrupt prompt and patches the hidden states at selected YES/NO marker token positions along a path from the corrupt relation-marker cloud toward the clean relation-marker cloud. Recovery is measured in two ways: whether the clean-minus-corrupt answer logit gap moves toward the clean-run value, and whether the residual relation geometry moves toward the clean run. In the representative pair used throughout this section, the clean answer describes the identity map, while the corrupt answer describes the duplicate-target map.

\begin{figure}[H]
\centering
\includegraphics[width=0.90\linewidth]{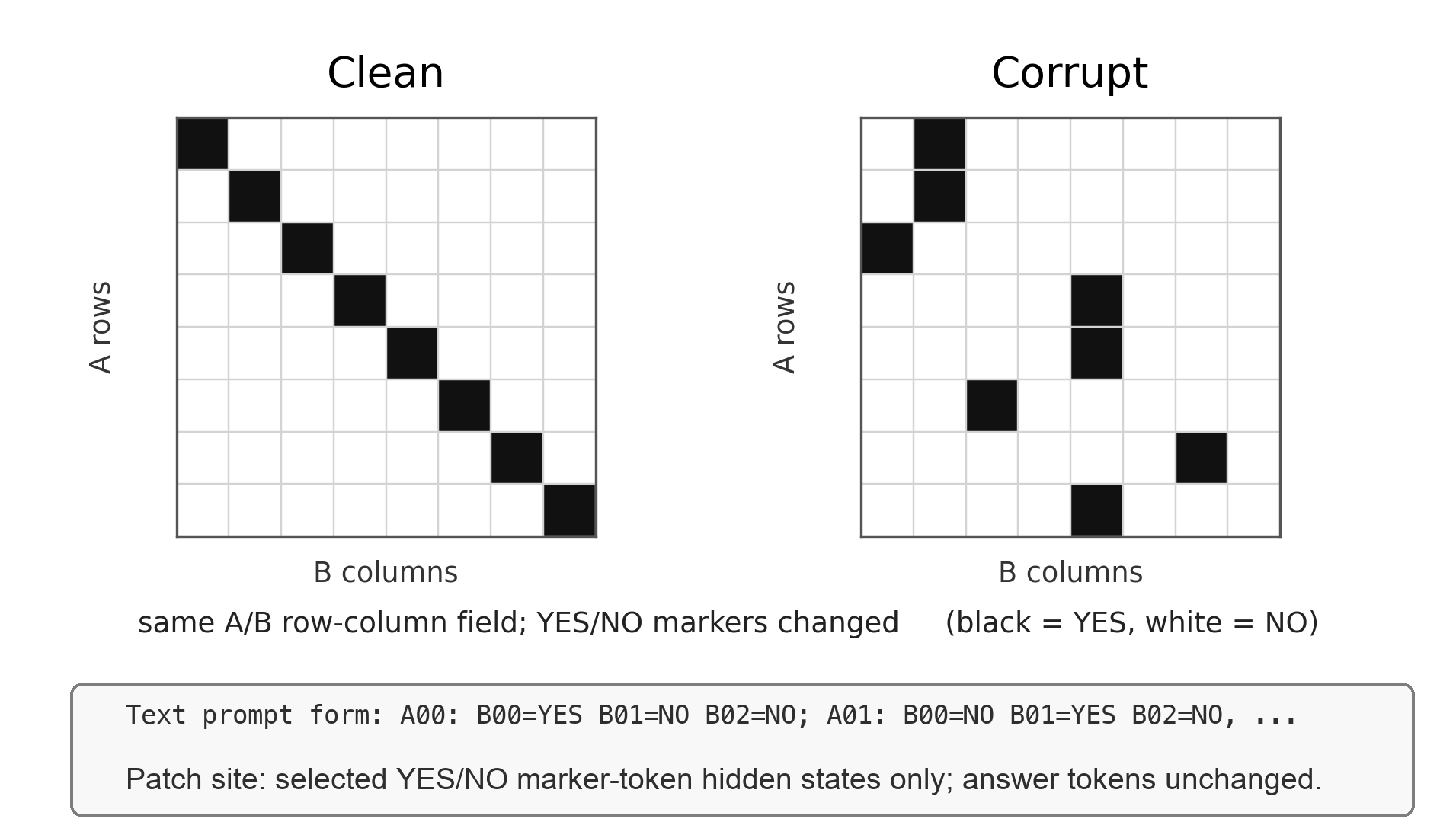}
\caption{Edge-grid clean/corrupt schematic. The row and column vocabulary stay fixed while the YES/NO relation markers change, creating a relation-specific answer flip and a localized set of marker token positions for intervention. The figure shows the same relation state as both a compact matrix and a literal prompt fragment.}
\label{fig:edge-schematic}
\end{figure}

\begin{table}[!t]
\centering
\small
\begin{tabular}{lccc}
\toprule
Model & clean accuracy & corrupt-answer selection accuracy & mean clean-corrupt logit gap \\
\midrule
8B & 0.250 & 0.250 & 0.029 \\
70B & 1.000 & 0.875 & 9.164 \\
405B & 1.000 & 1.000 & 14.609 \\
\bottomrule
\end{tabular}
\caption{Edge-grid baseline competence before steering. 405B is competent on both clean and corrupt sides; 70B is fully competent on the clean side and competent but not perfect on the corrupt side, with corrupt-answer selection $0.875$. 8B does not meet the competence threshold. Thus, the steering claim is made for 70B and 405B, with 8B treated as a competence-boundary run.}
\label{tab:edge-baseline}
\end{table}

\begin{figure}[!t]
\centering
\includegraphics[width=0.82\linewidth]{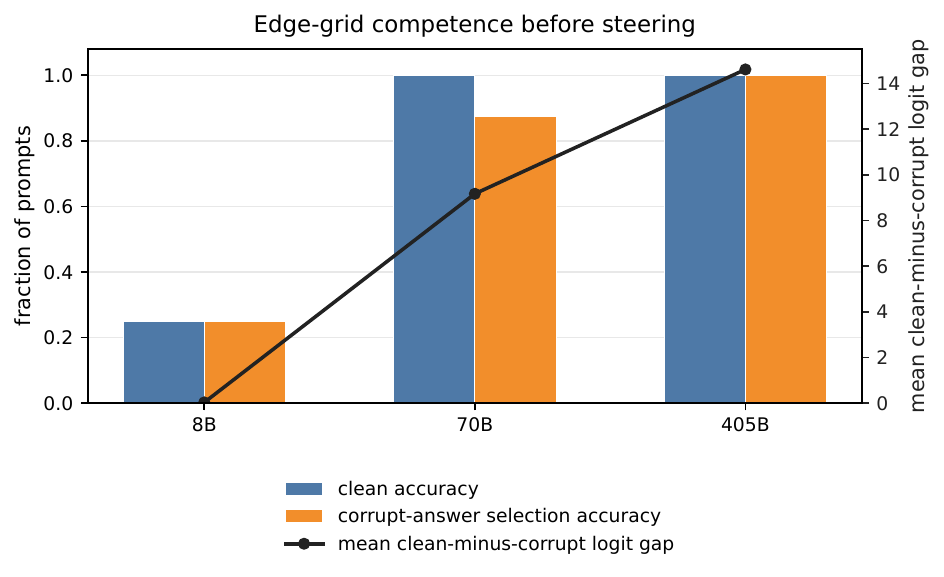}
\caption{Edge-grid baseline metrics before steering. Bars use the left axis and show clean accuracy and corrupt-answer selection accuracy; the black line uses the right axis and shows the mean clean-minus-corrupt answer logit gap.}
\label{fig:edge-baseline}
\end{figure}

With this diagnostic framework in place, the next section turns to relation-frame steering.

\section{Relation-Frame Steering}

The steering experiments are motivated by two constraints from recent interpretability work. First, internal structure is not enough: strong probes or internal representations can fail to translate into reliable output correction, producing a knowledge-action gap \citep{basu2026actionability}. Second, when internal variables are geometric objects rather than single directions, intervention paths should respect that geometry. Manifold-steering work shows that geometry-aware paths through activation space can induce more faithful behavioral trajectories than Euclidean straight-line steering \citep{wurgaft2026manifoldsteering,sarfati2026beliefs}. The edge-grid assay tests an analogous question for relations: if the clean and corrupt relation states define two relation-marker clouds, does moving the corrupt cloud along structured relation-frame paths recover both answer behavior and residual relation geometry?

\subsection{Intervention families}

All steering runs start from the corrupt prompt and patch relation-marker hidden states. The main 70B and 405B comparisons patch layer 5, measure residual relation geometry at layer 35, and measure final answer behavior from the output logits, using fixed layers across all intervention families. The patch is applied at layer 5, after which the forward pass continues normally to the readout layer. Each path is evaluated at $\alpha=0,0.05,\ldots,1.0$, with 32 prompts, 19 evaluated method rows, projection dimension $64$, projection seed $42$, subspace dimension $8$, tuple budget $8$, and 300 bootstrap resamples. One of those rows is an alias: in this assay, the linear marker path and centroid-plus-shape path induce the same sampled intervention, leaving 18 independent path constructions.

Operationally, this gives five intervention families: direct relation-frame motion, shape motion, centroid motion, discrete edge edits, and matched controls. A centroid method moves the mean of the marker cloud. A shape method changes the relative arrangement of markers after the mean has been removed. A Procrustes method uses the best rigid rotation aligning one centered cloud to another. A Grassmann method treats the subspace spanned by the centered cloud as the object being moved, so the path is between relation-shape subspaces rather than between single vectors.

\begin{itemize}
\item \textbf{Linear marker path:} add $\alpha$ times the clean-minus-corrupt marker-token delta to the corrupt relation-marker cloud.
\item \textbf{Shape-only:} keep the corrupt centroid fixed, but replace the relative arrangement of marker tokens with the clean centered shape.
\item \textbf{Centroid-plus-shape:} move the full corrupt relation cloud to the clean relation cloud.
\item \textbf{Procrustes and Grassmann shape paths:} align the centered marker cloud by a rigid rotation, or move the subspace spanned by that centered cloud toward the clean subspace, rather than only translating the mean.
\item \textbf{Discrete edge-dose and Hamming paths:} move a subset of edge markers, or a discrete intermediate edge map, toward the clean relation.
\item \textbf{Centroid-only:} translate the corrupt relation cloud by the clean-minus-corrupt centroid shift.
\item \textbf{Random controls and equal-norm noise:} match aspects of norm, centroid size, rotation, or subspace angle without using the clean relation structure.
\end{itemize}

Unlike manifold steering, these paths do not fit a continuous conceptual manifold over many semantic states. They use a paired clean/corrupt relation state and ask which components of the finite relation-marker frame matter, such as the centroid, centered shape, Procrustes alignment, or Grassmann subspace. This makes the intervention more local and task-specific, but also gives sharper controls over whether recovery follows the relation-frame shape rather than perturbation size or mean translation.

I also run a compact site-and-ordering audit using the same patch layer, readout layer, prompt count, and path fractions. This audit keeps the same selected YES/NO marker patch sites while breaking the ordered clean frame in several ways: permuting the clean centered marker rows, reflecting the clean shape, importing a clean centered shape from another prompt, importing a corrupt centered shape from another prompt, or applying the clean relation delta to matched non-target sites. The reported site-and-ordering summaries use the same 32 prompts and 500 prompt-level bootstrap resamples where bootstrap intervals are computed; the corrupt-donor rows are matched 70B/405B controls.

\subsection{Measuring recovery along a path}

Let $g_{i,\clean}$ be the clean-prompt logit difference between the clean answer and corrupt answer for prompt $i$, let $g_{i,\corr}$ be the corresponding corrupt-prompt value, and let $g_{i,\patch}(\alpha)$ be the patched corrupt-prompt value at path fraction $\alpha$. The behavior recovery is
\begin{equation}
R^{\mathrm{beh}}_i(\alpha)=\frac{g_{i,\patch}(\alpha)-g_{i,\corr}}{g_{i,\clean}-g_{i,\corr}}.
\end{equation}
Thus $0$ means no recovery from the corrupt baseline, and $1$ means the patched corrupt prompt recovers the clean-prompt answer gap. Values slightly above $1$ indicate that the patched corrupt prompt exceeds the clean-prompt clean-minus-corrupt logit gap under this normalization.

Geometry recovery asks whether the row-column planes associated with changed edges move from the corrupt relation state toward the clean relation state. For geometry, the run constructs a residual relation-blade readout vector at the readout layer. Hidden states are projected to the same $64$-dimensional analysis subspace. For each changed edge row, the code forms normalized exterior two-blades $x_a\wedge x_b$ from the row token and candidate column token, combines them into a local rectangle contrast for the clean-versus-corrupt target pair, normalizes that contrast, and averages these contrasts across changed rows. In simpler terms, the readout compares the local two-dimensional plane spanned by a row token and a candidate column token, rather than only comparing individual token vectors. This produces a unit vector summarizing the local residual relation geometry for the prompt.

Let $v_{i,\clean}$, $v_{i,\corr}$, and $v_{i,\patch}(\alpha)$ be the clean, corrupt, and patched residual relation vectors. The residual blade recovery is
\begin{equation}
R^{\mathrm{res}}_i(\alpha)=\frac{\cos(v_{i,\patch}(\alpha),v_{i,\clean})-\cos(v_{i,\corr},v_{i,\clean})}{1-\cos(v_{i,\corr},v_{i,\clean})}.
\end{equation}
The pointwise coupled recovery score is
\begin{equation}
R^{\mathrm{coup}}_i(\alpha)=R^{\mathrm{beh}}_i(\alpha)R^{\mathrm{res}}_i(\alpha),
\qquad
\mathrm{AUC}^{\mathrm{coup}}_i
=\sum_{j=1}^{m}(\alpha_j-\alpha_{j-1})
\frac{R^{\mathrm{coup}}_i(\alpha_j)+R^{\mathrm{coup}}_i(\alpha_{j-1})}{2}.
\end{equation}
I use the product because a path should receive high coupled recovery only when behavior and residual geometry move together at the same path fraction; isolated movement in only one readout remains small. After computing these quantities at each sampled path fraction, I summarize each intervention path at the endpoint and along the path. Endpoint behavior recovery and endpoint residual recovery measure whether the final patched state reaches the clean answer and clean residual geometry. Behavior-geometry correlation and coupled AUC measure whether these two recoveries move together across intermediate path fractions. The correlation column is the Pearson correlation between prompt-averaged behavior and residual recovery across the sampled path fractions. Off-target answer AUC checks whether the intervention also increases probability on unrelated answer choices. Off-target answer AUC is computed over the non-clean, non-corrupt answer options along the same path fractions, so small values indicate that recovery is not driven by broad probability leakage into unrelated choices. Confidence intervals in the main steering table are prompt-bootstrap 95\% intervals over 300 bootstrap resamples.

\subsection{Relation-frame paths recover behavior and geometry}

Table~\ref{tab:main-steering} reports representative steering families for the two behaviorally competent models, 70B and 405B. The comparison separates three outcomes: endpoint recovery, behavior-geometry coupling along the path, and off-target answer drift.

Smooth relation-frame paths reach near-complete endpoint recovery in both models. Linear marker, centroid-plus-shape, shape-only, and Grassmann shape recover the clean answer and residual geometry with small off-target AUC. In this assay, the linear marker path and the centroid-plus-shape path coincide as sampled implementations: both add the full clean-minus-corrupt marker delta at fraction $\alpha$, so they are aliases of one another rather than independent paths. Their rows in Tables~\ref{tab:main-steering},~\ref{tab:70b-full}, and~\ref{tab:405b-full} are therefore reported as duplicates and should be read as a single intervention. Discrete edge-dose paths reach strong endpoints but lower coupled AUC; random rotation is partial; centroid-only/equal-norm controls do not recover behavior. The main effect therefore follows structured movement of the relation frame rather than perturbation scale or centroid translation alone.

\begin{table}[H]
\centering
\scriptsize
\setlength{\tabcolsep}{4.7pt}
\renewcommand{\arraystretch}{1.24}
\makebox[\linewidth][c]{%
\resizebox{1.08\linewidth}{!}{%
\begin{tabular}{llccccc}
\toprule
Model & method & endpoint beh. & endpoint res. & corr. & coupled AUC & off-target AUC \\
\midrule
70B & linear marker & 1.006 [0.991, 1.019] & 0.895 [0.887, 0.902] & 0.863 & 0.378 [0.367, 0.388] & 0.007 \\
70B & centroid+shape (alias) & 1.006 [0.991, 1.019] & 0.895 [0.887, 0.902] & 0.863 & 0.378 [0.368, 0.388] & 0.007 \\
70B & shape only & 1.006 [0.992, 1.021] & 0.895 [0.888, 0.902] & 0.860 & 0.375 [0.364, 0.385] & 0.007 \\
70B & Grassmann shape & 1.006 [0.994, 1.019] & 0.895 [0.888, 0.902] & 0.860 & 0.373 [0.361, 0.382] & 0.007 \\
70B & edge dose & 1.006 [0.992, 1.019] & 0.895 [0.889, 0.903] & 0.775 & 0.211 [0.192, 0.228] & 0.006 \\
70B & random rotation & 0.478 [0.361, 0.589] & 0.742 [0.688, 0.790] & 0.861 & 0.073 [0.051, 0.098] & 0.009 \\
70B & centroid only & 0.002 [-0.003, 0.006] & 0.001 [-0.001, 0.002] & -0.010 & 0.000 [-0.000, 0.000] & 0.008 \\
70B & equal-norm noise & 0.024 [0.012, 0.037] & 0.016 [0.006, 0.024] & 0.146 & 0.000 [-0.000, 0.000] & 0.008 \\
\midrule
405B & linear marker & 0.996 [0.992, 1.000] & 0.882 [0.862, 0.902] & 0.976 & 0.365 [0.352, 0.376] & 0.001 \\
405B & centroid+shape (alias) & 0.996 [0.992, 1.000] & 0.882 [0.862, 0.901] & 0.976 & 0.365 [0.352, 0.376] & 0.001 \\
405B & shape only & 0.995 [0.991, 1.000] & 0.881 [0.861, 0.898] & 0.975 & 0.362 [0.351, 0.374] & 0.001 \\
405B & Grassmann shape & 0.995 [0.991, 1.000] & 0.881 [0.859, 0.900] & 0.974 & 0.361 [0.349, 0.371] & 0.001 \\
405B & edge dose & 0.996 [0.992, 1.000] & 0.882 [0.862, 0.900] & 0.808 & 0.181 [0.167, 0.197] & 0.001 \\
405B & random rotation & 0.603 [0.480, 0.711] & 0.570 [0.495, 0.649] & 0.851 & 0.068 [0.045, 0.091] & 0.001 \\
405B & centroid only & 0.003 [-0.001, 0.007] & -0.000 [-0.002, 0.001] & -0.011 & 0.000 [0.000, 0.000] & 0.001 \\
405B & equal-norm noise & 0.023 [0.014, 0.033] & -0.027 [-0.043, -0.011] & -0.296 & -0.000 [-0.000, 0.000] & 0.001 \\
\bottomrule
\end{tabular}
}
}
\caption{Main steering comparison. Endpoint behavior recovery measures movement toward the clean-answer logit gap, while endpoint residual recovery measures movement toward the clean residual relation geometry at readout. Coupled AUC summarizes how behavior and residual recovery move together across path fractions. Bootstrap intervals are shown for endpoint behavior recovery, endpoint residual recovery, and coupled AUC; correlation and off-target AUC are reported as point estimates. Smooth relation-frame paths recover behavior and residual geometry in 70B and 405B. Centroid-only and equal-norm noise do not. The linear marker and centroid+shape rows are alias implementations of the same sampled path in this assay; identical values are expected and not an additional independent success.}
\label{tab:main-steering}
\end{table}

\begin{figure}[!t]
\centering
\makebox[\linewidth][c]{\includegraphics[width=0.96\linewidth]{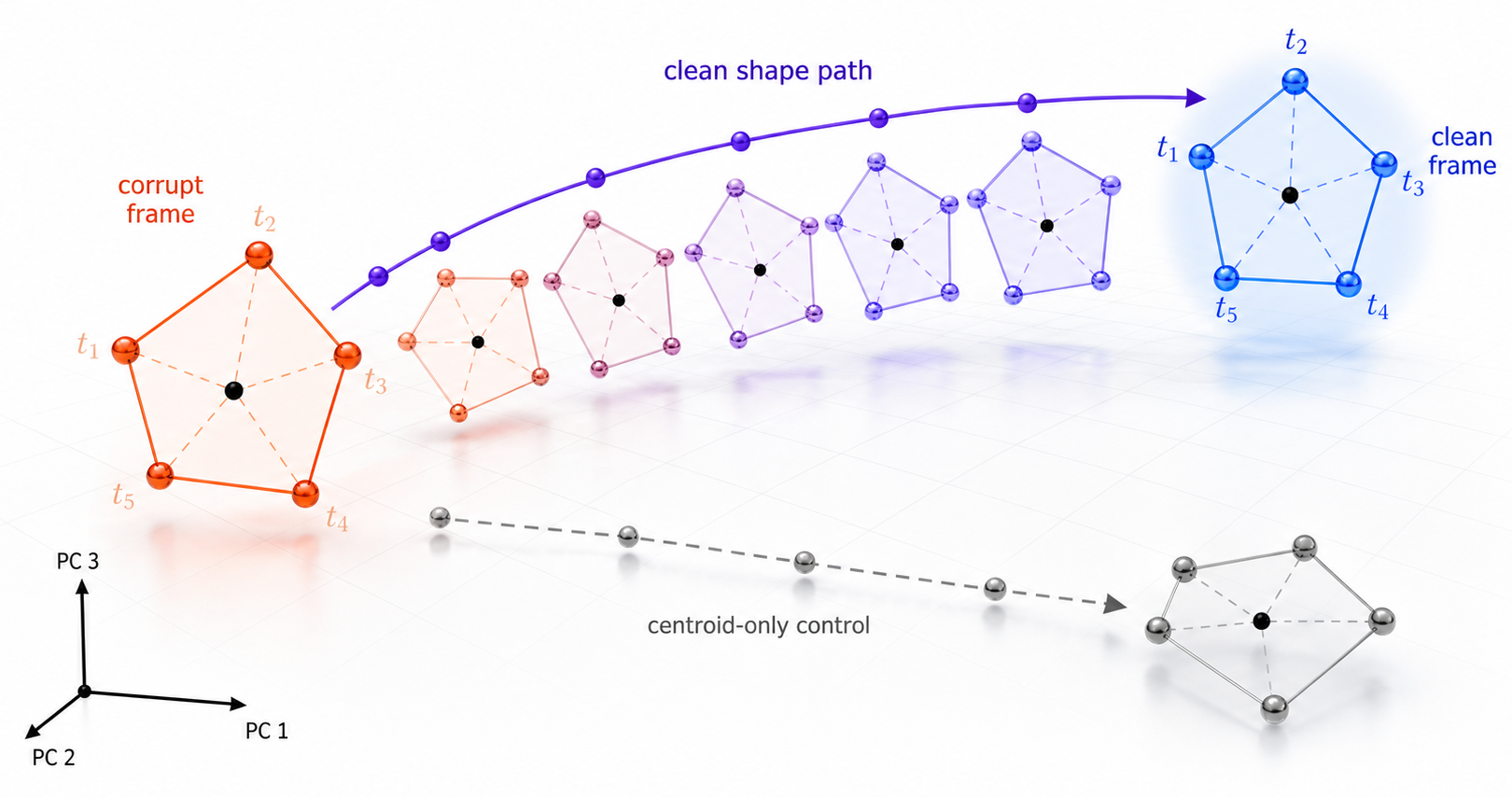}}
\caption{Relation-frame shape path schematic. Orange and blue clouds show corrupt and clean relation-marker frames in a low-dimensional projection. The purple path depicts the clean-targeted shape intervention, while the gray dashed path shows the centroid-only control that moves the mean without aligning the relative shape. Quantitative results are in Table~\ref{tab:main-steering} and Figure~\ref{fig:method-auc}.}
\label{fig:shape-path-schematic}
\end{figure}

\subsection{Separating shape from centroid motion}

The strongest mechanistic contrast is between structured relation-marker cloud shape and mean translation. Centroid-only translation has almost no effect: endpoint behavior recovery is $0.002$ in 70B and $0.003$ in 405B. Equal-norm noise also stays negligible, with endpoint behavior recovery of $0.024$ in 70B and $0.023$ in 405B; in 405B, residual blade recovery is slightly negative. In contrast, shape-only replacement nearly matches full centroid-plus-shape replacement. Recovery therefore does not stem from translating the mean hidden state of the relation-marker cloud, nor from applying a matched-size perturbation. The causally effective object is the structured shape of the relation-marker cloud.

\subsection{Site/order controls and clean-state transfer}

The main steering run patches selected YES/NO marker-token positions, so the site-and-ordering audit separates the contribution of those sites from the ordered clean-frame geometry. It preserves the patching interface and marker sites while breaking order, orientation, donor clean state, or target site.

Table~\ref{tab:site-order-controls} gives the completed 70B audit together with the matched 405B audit. Ordered clean shape and cross-prompt clean shape both recover behavior and residual geometry in both models. In contrast, the same marker sites are not sufficient. Permuting the clean centered rows or reflecting the shape sharply weakens or abolishes behavior recovery, and the 70B/405B corrupt-donor same-site shapes stay near zero in coupled recovery. Because this table is a separately run site-and-ordering control batch rather than a rowwise duplicate of Table~\ref{tab:main-steering}, shared method names such as shape-only are calibration rows inside the audit and endpoint residual values should be compared within each audit block. Edge-marker Pl\"ucker recovery is normalized by the clean-minus-corrupt contrast, so values above $1$, such as the \(1.348\) 70B shape-only value in Table~\ref{tab:site-order-controls}, indicate overshoot beyond the clean-state scalar readout rather than an invalid recovery score.

\begin{table}[t]
\centering
{\scriptsize
\setlength{\tabcolsep}{4pt}
\renewcommand{\arraystretch}{1.12}
\resizebox{0.95\linewidth}{!}{%
\begin{tabular}{llrrrrr}
\toprule
Model & method & endpoint beh. & endpoint res. & edge rec. & coupled AUC & off-target AUC \\
\midrule
\multicolumn{7}{l}{\textit{70B target and transfer paths}} \\
70B & shape only & 1.006 & 0.719 & 1.348 & 0.301 & 0.0067 \\
70B & cross-prompt clean shape & 1.009 & 0.718 & 1.078 & 0.302 & 0.0068 \\
70B & cross-prompt corrupt shape & 0.024 & 0.003 & 0.018 & 0.000 & 0.0055 \\
\multicolumn{7}{l}{\textit{70B order/site/scale placebos}} \\
70B & permuted same-site shape & 0.140 & 0.309 & 0.551 & 0.023 & 0.0081 \\
70B & reflected same-site shape & 0.004 & -0.032 & 0.096 & 0.000 & 0.0074 \\
70B & wrong-site clean delta & -0.096 & -0.003 & -0.310 & 0.001 & 0.0072 \\
70B & centroid only & 0.002 & 0.011 & -0.197 & 0.000 & 0.0075 \\
70B & equal-norm noise & 0.024 & 0.006 & 0.256 & 0.000 & 0.0077 \\
\midrule
\multicolumn{7}{l}{\textit{405B target and transfer paths}} \\
405B & shape only & 0.991 & 0.852 & 0.992 & 0.358 & 0.0002 \\
405B & cross-prompt clean shape & 0.992 & 0.847 & 1.009 & 0.365 & 0.0002 \\
405B & cross-prompt corrupt shape & -0.001 & -0.002 & 0.078 & 0.000 & 0.0002 \\
\multicolumn{7}{l}{\textit{405B order/site/scale placebos}} \\
405B & permuted same-site shape & 0.093 & 0.419 & 0.659 & 0.026 & 0.0003 \\
405B & reflected same-site shape & 0.005 & -0.004 & 0.170 & -0.000 & 0.0002 \\
405B & wrong-site clean delta & -0.040 & 0.016 & 0.434 & 0.000 & 0.0004 \\
405B & centroid only & -0.004 & -0.001 & -0.008 & -0.000 & 0.0002 \\
405B & equal-norm noise & 0.019 & -0.029 & 0.093 & -0.000 & 0.0002 \\
\bottomrule
\end{tabular}
}
}
\caption{Site/order controls and clean-state transfer. The table separates successful clean-state shape paths from controls that preserve the patch interface while breaking order, orientation, donor clean state, or target site. Both audits use 32 prompts, patch layer 5, readout layer 35, and path fractions $0,0.05,\ldots,1.0$. Because this is a separately computed site/order control batch, shared method names are audit calibration rows and endpoint residual values should be compared within this table rather than row-by-row to Table~\ref{tab:main-steering}. Edge rec. is normalized edge-marker Pl\"ucker recovery; values above $1$ indicate overshoot of the clean reference. The corrupt-donor rows are matched controls.}
\label{tab:site-order-controls}
\end{table}

The opposite dissociation also holds, since applying the clean relation delta to matched non-target sites fails. Successful recovery depends on a clean-state relation frame placed at the marker interface, not merely on patching marker-token positions or injecting a matched perturbation. The cross-prompt clean-shape result has a specific scope: in this prompt bank, the clean edge map is the identity map for every prompt, while corrupt maps and answer labels vary. Within that scope, cross-prompt success supports a portable clean/solved identity-frame format inside this edge-grid family, but does not by itself prove transfer across arbitrary clean relation maps. Testing arbitrary clean relation-map transfer would require varying the clean target map itself and asking whether donor clean frames transfer across genuinely different target relations. The matched 70B/405B corrupt-donor failures sharpen the interpretation, showing that the portable object is clean-state specific rather than a generic donor marker-cloud activation.

\begin{figure}[!t]
\centering
\makebox[\linewidth][c]{\includegraphics[width=1.18\linewidth]{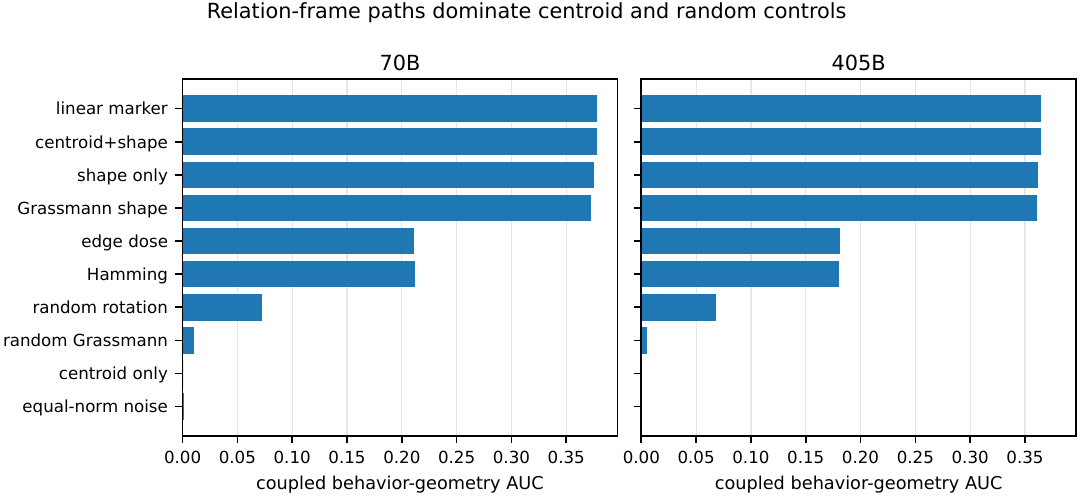}}
\caption{Coupled behavior--geometry AUC by steering method. Coupled AUC integrates simultaneous clean-answer behavior recovery and residual-geometry recovery over the sampled path fractions. Direct and shape-based relation-frame paths have the largest coupled AUC, while discrete paths and random controls are weaker along the path.}
\label{fig:method-auc}
\end{figure}

\subsection{Smooth paths versus discrete edge-dose paths}

Discrete edge-dose and Hamming paths also reach strong endpoints, but with lower coupled AUC. In 70B, edge-dose coupled AUC is $0.211$ versus $0.378$ for the linear marker path; in 405B, it is $0.181$ versus $0.365$. This separates endpoint recovery from path quality: discrete paths can arrive at a clean endpoint without coupling behavior and residual geometry as strongly along the route. The smoother relation-frame paths better capture the continuous intervention geometry than a count of corrected edge markers.

\begin{figure}[H]
\centering
\makebox[\linewidth][c]{\includegraphics[width=1.12\linewidth]{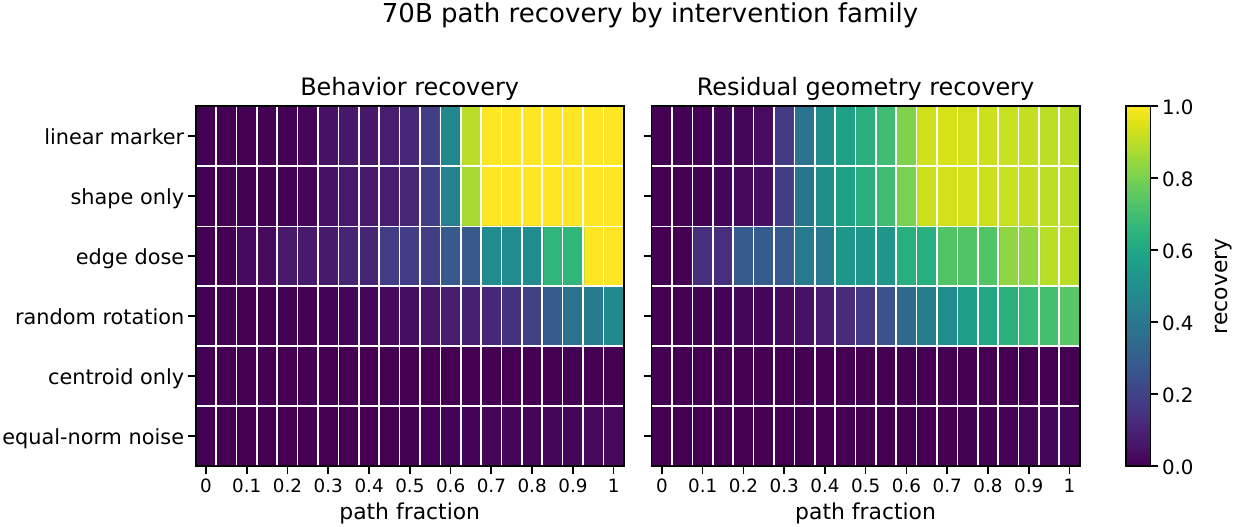}}
\caption{70B path recovery heatmaps across sampled path fractions. Each heatmap traces how behavior recovery and residual-geometry recovery change as the patched relation-marker cloud moves from corrupt toward clean. Linear marker and shape-only paths show smooth recovery in both quantities. Edge-dose recovers in step-like bands, random rotation is partial, and centroid-only/equal-norm controls do not recover.}
\label{fig:path-70b}
\end{figure}

\begin{figure}[!t]
\centering
\makebox[\linewidth][c]{\includegraphics[width=1.03\linewidth]{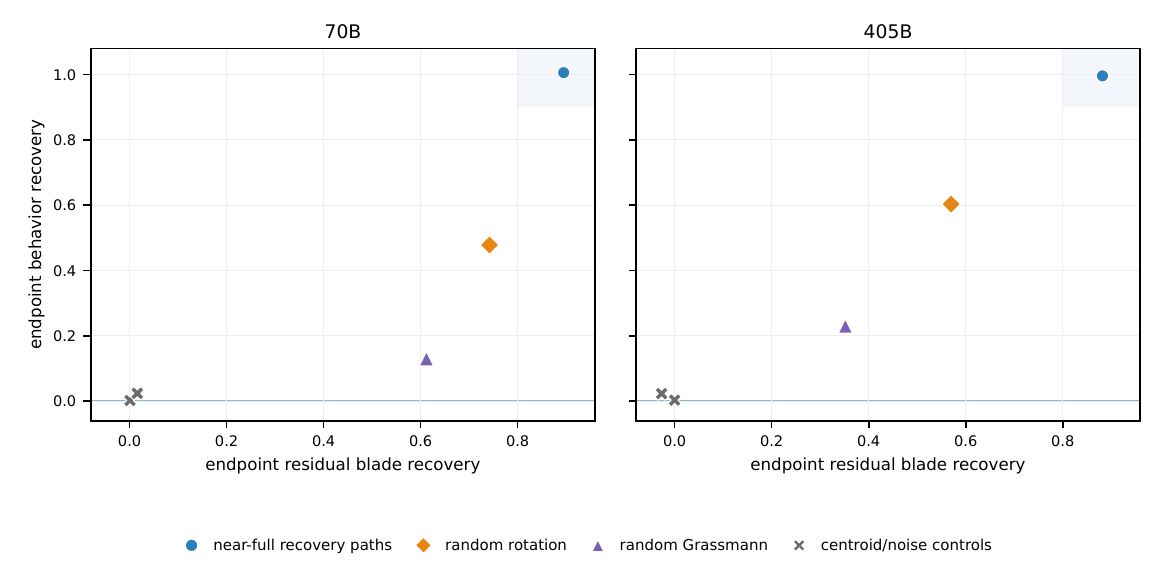}}
\caption{Endpoint recovery frontier. The horizontal axis is endpoint residual-geometry recovery and the vertical axis is endpoint clean-answer behavior recovery. Point styles group overlapping methods for readability. Near-full recovery paths occupy the upper-right frontier; random rotation and random Grassmann paths are partial controls; centroid/noise controls cluster near the origin.}
\label{fig:endpoint-frontier}
\end{figure}

\subsection{Pl\"ucker readouts along steering paths}

Geometry enters the paper in two connected roles. Pl\"ucker sign entropy first serves as a diagnostic, asking whether relation tuples have consistent determinant-sign structure at the expected rank. Relation-frame steering later serves as an intervention, moving a selected cloud of relation-marker hidden states from a corrupt relation state toward a clean one. The bridge is not that the diagnostic and steering banks use identical token sets; it is that both operate on ordered relation-frame clouds. The edge-grid steering runs also record an edge-grid Pl\"ucker readout along the same patch paths.

For this bridge, I use the statistic \(D^{2,3}_{\mathrm{edge}}\). The superscript \(2,3\) denotes the average of rank-2 and rank-3 local Pl\"ucker contrasts over edge-marker tuples. This is not a rerun of the full controlled-arity bank inside the grid; it is the same determinant-sign idea applied locally to the YES/NO marker tuples that define the edge-grid relation state. Formally, let \(E\) denote the selected YES/NO edge-marker token positions in an edge-grid prompt. For each \(k\in\{2,3\}\), ordered edge-marker tuple sets are enumerated from \(E\) in the fixed edge role order, with matched scrambled and random controls drawn from the same prompt under the same selection rule used for the diagnostic. Determinant signs are computed from the row-selected matrix \(U_k[t,:]\) of the projected hidden-state SVD, as in Section~2. Let
\[
D_{\mathrm{edge}}^{(k)}=\Hpm(T_{\mathrm{edge,scrambled}}^{(k)})-\Hpm(T_{\mathrm{edge,true}}^{(k)}),
\qquad
D_{\mathrm{edge}}^{2,3}=\frac{1}{2}\left(D_{\mathrm{edge}}^{(2)}+D_{\mathrm{edge}}^{(3)}\right).
\]
Matched random edge tuples are retained as audit controls, but the reported \(D^{2,3}_{\mathrm{edge}}\) recovery scalar uses the true-vs-scrambled contrast, matching the form used for \(D(r,k)\) after the random term cancels.
For a patched state at path fraction \(\alpha\), edge-Pl\"ucker recovery is the normalized corrupt-to-clean movement of this scalar,
\[
R_{\mathrm{edge}}(\alpha)=
\frac{D_{\mathrm{edge,patch}}^{2,3}(\alpha)-D_{\mathrm{edge,corrupt}}^{2,3}}
{D_{\mathrm{edge,clean}}^{2,3}-D_{\mathrm{edge,corrupt}}^{2,3}}.
\]
If the clean prompt has a characteristic orientation pattern over its relation markers, then a successful patch should not only move the answer logits. It should also move that marker-token orientation pattern back toward the clean value. The readout is measured after the intervention; the patch paths are not optimized for it.

Successful clean-targeted paths give the clearest bridge result. In 70B and 405B, linear, shape-only, and Grassmann shape paths recover behavior and residual blade geometry while also moving the edge-marker Pl\"ucker readout toward the clean state. This matters because the bridge readout tests the same kind of geometric structure used by the diagnostic. Recovery therefore means that the successful patch restores part of the clean prompt's oriented relation geometry, not just its output behavior. Centroid-only translation is negligible, showing that merely shifting the mean marker vector does not repair this orientation structure. Equal-norm noise also has negligible coupled edge-Pl\"ucker scores, ruling out a generic perturbation-size explanation. Random rotation remains intermediate, moving some endpoint readouts without producing the same coupled recovery as the targeted relation-frame paths. Successful interventions depend on moving the relation frame in a clean-aligned direction rather than simply changing its subspace geometry. Because edge-Pl\"ucker recovery is normalized from the corrupt value to the clean value, values above \(1\) indicate overshoot past the clean reference along the same scalar readout rather than amplified recovery.

\begin{table}[H]
\centering
{\scriptsize
\setlength{\tabcolsep}{2.6pt}
\renewcommand{\arraystretch}{0.95}
\resizebox{0.82\linewidth}{!}{
\begin{tabular}{lrrrrrrrr}
\toprule
& \multicolumn{4}{c}{70B} & \multicolumn{4}{c}{405B} \\
\cmidrule(lr){2-5}\cmidrule(lr){6-9}
method & beh. & res. & edge rec. & edge score & beh. & res. & edge rec. & edge score \\
\midrule
linear marker & 1.006 & 0.895 & 0.449 & 0.152 & 0.996 & 0.882 & 0.940 & 0.405 \\
shape only & 1.006 & 0.895 & 0.560 & 0.153 & 0.995 & 0.881 & 0.940 & 0.405 \\
Grassmann shape & 1.006 & 0.895 & 0.560 & 0.152 & 0.995 & 0.881 & 0.940 & 0.408 \\
random rotation & 0.478 & 0.742 & 0.387 & 0.040 & 0.603 & 0.570 & 0.788 & 0.098 \\
centroid only & 0.002 & 0.001 & 0.034 & 0.000 & 0.003 & -0.000 & -0.024 & -0.000 \\
equal-norm noise & 0.024 & 0.016 & 0.031 & 0.000 & 0.023 & -0.027 & 0.324 & -0.000 \\
\bottomrule
\end{tabular}
}
}
\caption{Edge-grid Pl\"ucker readouts along steering paths. Edge-Pl\"ucker recovery is normalized \(D^{2,3}_{\mathrm{edge}}\) movement, where \(D^{2,3}_{\mathrm{edge}}\) averages the rank-2 and rank-3 determinant-sign contrasts over local edge-marker tuples in the clean, corrupt, and patched states. Coupled edge-Pl\"ucker is the path-average behavior--\(D^{2,3}_{\mathrm{edge}}\) score. The normalization sets the corrupt state to \(0\) and the clean state to \(1\); values above \(1\) indicate overshoot of the clean reference under this scalar readout.}
\label{tab:edge-plucker-bridge}
\end{table}

\section{Controls and Scope}

This section delimits the two main claims of the paper. For the diagnostic claim, the controls ask whether rank-indexed structure survives template variation without treating the expected rank as the only active signal. For the steering claim, they ask whether recovery is specific to targeted relation-frame movement rather than task failure, centroid translation, perturbation size, random subspace motion, or broad answer destabilization.

\paragraph{Competence boundary.} The 8B model is not behaviorally competent on the edge-grid assay: clean accuracy is $0.250$, corrupt answer selection is $0.250$, and the mean clean-corrupt logit gap is $0.029$. Some 8B path metrics show geometry movement, but the answer behavior is not anchored by a reliable clean/corrupt behavioral task. 8B is therefore a competence-boundary run rather than a main steering success.

\paragraph{Centroid controls.} Centroid-only and random-centroid shifts produce negligible endpoint recovery in 70B and 405B. This rules out a pure mean-translation explanation for the successful interventions.

\paragraph{Energy and norm controls.} Equal-norm noise produces negligible endpoint behavior even though it matches the norm scale of the clean-minus-corrupt delta. Perturbation size alone is not sufficient to recover the clean answer.

\paragraph{Site/order controls and clean-state transfer.} The site-and-ordering audit tests the stronger possibility that selected YES/NO marker sites are sufficient even without the ordered clean frame. They are not sufficient. Same-site permutation, same-site reflection, wrong-site clean deltas, centroid-only translation, and equal-norm noise all fail behavior--geometry coupled recovery, while ordered clean shape and cross-prompt clean shape succeed. The matched 70B/405B corrupt-donor rows also fail, showing that cross-prompt transfer is clean-state specific rather than a generic donor-marker activation.

\paragraph{Cross-prompt transfer scope.} Cross-prompt clean-shape transfer shows that the clean/solved identity-frame format is portable inside the current edge-grid family. The boundary is precise: the audited clean map is always the identity map, so the result does not establish transfer across arbitrary relation contents. The 70B/405B corrupt-donor controls matter because they show that the donor cloud must be a clean-state cloud, not merely any donor marker cloud.

\paragraph{Random subspace controls.} Random rotations and random Grassmann paths produce partial movement in some metrics but much lower coupled AUC than targeted relation-frame paths. They serve as partial negative controls: they show that generic subspace-like movement can produce nonzero effects, but still falls far short of the clean relation-frame path. The high correlations for some random-rotation paths reflect coherent but incomplete movement, not full recovery.

\paragraph{Off-target behavior.} Off-target answer AUC stays very small in 70B and 405B, indicating that the main interventions move probability toward the clean relation answer rather than broadly destabilizing the answer distribution.

\paragraph{Template and rank controls.} Multi-template audits preserve positive diagonal rank structure, especially in 405B. The mixed 8B and 70B constructor-specific rows delimit the diagnostic claim.

The resulting scope is intentionally narrow: the diagnostic is rank- and constructor-conditioned, and the steering claim is made only for competent clean/corrupt edge-grid runs with matched controls.

\section{Related Work}

\subsection{Circuits, patching, and actionability}

Intervention-based mechanistic interpretability is the closest methodological precedent for the causal half of this paper. Transformer-circuit work decomposes behavior into attention heads, MLPs, residual-stream pathways, and circuit subgraphs \citep{elhage2021framework,wang2023interpretability,conmy2023acdc}. Activation patching, causal tracing, causal abstraction, and attribution patching ask whether internal states mediate behavior under clean/corrupt substitutions \citep{meng2022locating,geiger2021causal,syed2023attribution}. Recent circuit-tracing work pushes this program further by constructing attribution graphs: prompt-local computational graphs over interpretable features, with perturbation experiments used to validate graph hypotheses \citep{ameisen2025circuittracing,lindsey2025biology}. The present paper uses the same broad clean/corrupt intervention logic, but the intervention object is not a single feature, component, or circuit edge. It is a relation frame: a selected hidden-state configuration of relation-marker tokens.

The distinction between detection and intervention is important. Work on interpretability without actionability shows that highly predictive internal representations may still fail to support reliable output correction under existing steering methods \citep{basu2026actionability}. This paper therefore pairs relation-frame detection with behavioral intervention. The edge-grid assay asks whether moving the selected relation frame from corrupt toward clean recovers both answer logits and residual relation geometry.

\subsection{Directions, sparse features, and representation manifolds}

Sparse autoencoders and linear representation methods search for interpretable features, directions, or subspaces in activation space \citep{bricken2023monosemanticity,cunningham2024sparse,park2024linear,hernandez2024linearity}. However, recent work argues that some model variables are better treated as geometric objects rather than isolated directions. Sparse autoencoders may capture concept manifolds globally, locally, or in a fragmented regime, motivating methods that search for coherent groups of atoms rather than single features \citep{bhalla2026saemanifolds}. Other work develops hypothesis-driven feature-manifold analysis, showing that temporal concepts can instantiate circles, lines, and clusters that support reasoning and reshape with context \citep{tiblias2026smds}. Belief-manifold and manifold-steering work similarly suggests that geometry-aware interventions can better preserve the intended behavioral family than globally linear steering \citep{sarfati2026beliefs,wurgaft2026manifoldsteering}.

The present paper follows this geometric turn, but studies a different object. A relation frame is not a continuous concept manifold such as a month circle or posterior surface. It is a finite ordered configuration of token hidden states selected by a relation. Pl\"ucker sign entropy measures the orientation structure of these configurations, while the steering assay tests whether moving the frame's shape changes relation-dependent behavior.

\subsection{Relational binding and relation geometry}

A growing literature studies how language models encode relations among entities. Work on relation decoding and entity binding shows that contextual entities and relations can be represented in hidden states \citep{hernandez2024linearity,feng2024binding}. Recent cell-based binding work finds a low-dimensional subspace in which cells correspond to entity--relation index pairs, with grid-like geometry and causal activation-patching effects on relational predictions \citep{dai2026cellbinding}. Separately, syntactic geometry work shows that syntactic relation type and direction can be read from distance and relative direction in low-dimensional activation subspaces \citep{diegosimon2024polar}. These results suggest that relations are not merely labels attached to individual tokens; they have organized geometric structure.

This paper extends that direction to higher-order ordered tuples. Instead of decoding relation labels or entity--relation indices, it tests whether relation-bound tuples have rank-indexed determinant-sign structure. The emphasis is therefore not only on whether a relation is linearly decodable, but on whether the participating tokens form an oriented hidden-state frame whose rank matches the relation arity.

\subsection{Relational complexity and higher-order structure}

Relational-complexity theory uses arity to distinguish unary, binary, ternary, and higher-order binding demands \citep{halford1998relational,gentner1983structure}. Recent REL evaluations show that model performance degrades as the number of simultaneously bound entities or operands increases, even under controls for input size and exposure to examples \citep{fesser2026rel}. This motivates an internal diagnostic for higher-order relation structure. The controlled-arity bank in this paper fixes the relation arity by construction and asks whether the hidden-state geometry shows an arity-matched orientation signature.

The mathematical language comes from Pl\"ucker coordinates, Grassmann geometry, and oriented matroids \citep{bjorner1999oriented}. Pl\"ucker coordinates represent subspaces through minors; Grassmann geometry treats subspaces as geometric objects; and oriented matroids retain determinant sign patterns after metric scale has been discarded. Here, determinant signs of selected token-tuple minors serve as an empirical statistic for relation-frame orientation. Higher-order networks and simplicial methods likewise emphasize structure beyond pairwise edges \citep{battiston2020networks,benson2018simplicial}, but here the higher-order object is a tuple of hidden-state token vectors inside a transformer.

Relation rank geometry sits inside this broader geometry/actionability turn, but at the level of relation-bound token tuples. Pl\"ucker sign entropy supplies a rank-indexed diagnostic, and the edge-grid experiments ask whether the associated relation frame is behaviorally actionable. The steering result does not replace circuit localization, but it provides a relation-level intervention object: a hidden-state frame whose structured movement recovers relation-dependent behavior.

\section{Limitations}

At the level of the claim, this paper studies relation frames as state-level objects. A relation frame is a selected hidden-state configuration of relation-bound tokens, used first for measurement and then for intervention. The claim is not that the full circuit has been localized. It leaves open which upstream heads or MLPs construct the frame, and which downstream components consume it. This scope is deliberate. Recent work on cyclic reasoning shows that visible representation geometry can coexist with a distinct computational mechanism \citep{feucht2026arithmeticwild}. The present results fit that division, with relation-frame geometry acting as a state object available to the model while the surrounding construction and readout circuit remains a separate target for analysis.

Empirically, the steering evidence is concentrated in the edge-grid family, where relation changes are explicit and the clean/corrupt answer flip is behaviorally anchored. The main steering claim is made for Llama-family 70B and 405B checkpoints over 32 prompts, patch layer 5, and readout layer 35. The 8B run is reported as a competence-boundary case rather than as primary causal evidence. The 70B and 405B runs use the 4-bit BitsAndBytes loading regime with bfloat16 compute documented in the reproducibility materials. All within-model controls, including the matched corrupt-donor controls, use the same regime, so comparisons across paths, sites, and controls are internally consistent.

A separate boundary concerns cross-prompt clean-shape transfer. The clean edge map is the identity map throughout the current prompt bank, while corrupt maps, prompt instances, and answer labels vary. The result therefore supports portability of a clean/solved identity-frame format inside this edge-grid family, and the matched corrupt-donor failures show that this portability is clean-state specific rather than a generic same-site activation effect. Transfer across genuinely different clean relation maps is a separate question that requires a clean-map-varied bank.

Positive expected-rank \(D\) values and positive constructor margins indicate that the arity-matched rank is enriched relative to the tested alternatives. They do not mean that lower ranks are inactive, that rank uniquely labels the computation, or that every constructor behaves uniformly. Lower-rank shadows, constructor-specific strength, and mixed cells in some 8B and 70B multi-template settings are part of the empirical picture. The controlled bank isolates arity by construction. In naturalistic prompts, arity may be harder to isolate because relational structure may be distributed across predicates, arguments, syntax, discourse context, and retrieval cues.

Mechanistically, the next step is to connect the state-level object studied here to circuit-level accounts of how it is built, routed, and read out. Circuit tracing or attribution-graph methods could identify the heads and MLP features that construct the relation frame at the patch layer and the downstream components that consume it at readout. A separate extension is to move beyond the edge-grid scaffold and test whether relation-frame steering transfers to less controlled naturalistic relation prompts, where the relevant relation state is not explicitly marked by a YES/NO grid. Other complementary extensions include relation banks with varied clean target maps, additional model families and architectures, pre-specified patch and readout layer sweeps, and full-precision replications of the large-model runs as loading regimes evolve. The paper isolates a measurable and steerable relation-state object, while future work can locate the machinery that constructs and uses that object.

\section{Reproducibility Materials}
\enlargethispage{2\baselineskip}

Code and reproducibility materials are available at GitHub: \url{https://github.com/mazenkobrosly/relational-rank-geometry}. The accompanying replication bundle contains the prompt banks, run configurations, steering scripts, row-level steering CSVs, path-quality tables, bootstrap summaries, multi-template outputs, and SHA256/file manifests used to audit the reported results. For the reported Llama-family diagnostics, the public checkpoint IDs are \texttt{meta-llama/Llama-3.1-8B-Instruct}, \texttt{meta-llama/Llama-3.1-70B-Instruct}, and \texttt{meta-llama/Llama-3.1-405B-Instruct}. The main steering numbers in Table~\ref{tab:main-steering} are drawn from \path{edge_grid_centroid_rotation_path_quality.csv} for the reported 70B and 405B runs. The steering run configs record local model paths \path{/workspace/models/llama31_8b_instruct}, \path{/workspace/models/llama31_70b_instruct}, and \path{/dev/shm/llama31_405b_official_split}. The site-and-ordering audit and matched corrupt-donor controls are recorded in the corresponding 70B and 405B run manifests. The reported 70B and 405B large-model runs use 4-bit BitsAndBytes model loading with bfloat16 compute, as recorded in the run configurations; cached activation arrays are stored in reduced precision according to the accompanying manifests. The 70B diagnostic rows use the corresponding 70B revision packet. The matched corrupt-donor steering outputs passed row-count, duplicate-row, and missing-field checks before inclusion. Model checkpoints are not redistributed. These materials make the reported tables traceable to concrete run artifacts while keeping model weights external.

\section{Conclusion}

The diagnostic and intervention experiments isolate a relation-level object inside transformer hidden states. Pl\"ucker sign entropy reads rank-indexed orientation structure from selected token tuples, and across Llama-family 8B, 70B, and 405B checkpoints the arity-matched rank $k=r$ is enriched relative to scrambled controls under matched random-control audits for $r=3,\ldots,6$. This structure is most uniform in 405B, where multi-template audits retain positive diagonal margins across argument-only and predicate-plus-argument constructors, although its raw expected-rank $D$ magnitudes are not uniformly the largest across models. At smaller scales it remains structured rather than uniform, with constructor-specific cells in 8B and 70B.

In behaviorally anchored 70B and 405B edge-grid runs, the corresponding relation frame becomes an intervention object rather than only a measurement. Moving the corrupt relation-marker cloud along clean-targeted shape, centroid-plus-shape, and Grassmann shape paths recovers clean-answer behavior and residual relation geometry, while centroid-only translation, equal-norm noise, same-site permutations and reflections, wrong-site clean deltas, and matched corrupt-donor frames remain far below. The causally effective handle is the ordered, clean-state shape of the relation-marker cloud at the patched marker sites. Cross-prompt clean-shape transfer succeeds within the current prompt bank, where the clean map is always the identity map, and the matched corrupt-donor failures show that this portability is specific to the clean-state frame rather than to the marker sites themselves.

The contribution is therefore a state-level one: a hidden-state configuration whose orientation structure can be measured by determinant signs and whose ordered shape can be moved to steer relation-dependent behavior. Future work can connect this object to the components that build and consume it through circuit-tracing or attribution-graph methods, clean-map-varied prompt banks, additional model families, and full-precision replications. Within those scopes, relational rank geometry is both detectable and steerable, giving mechanistic interpretability a concrete relation-level handle alongside feature-, direction-, and circuit-level objects.

\appendix

\section{Controlled-Arity Held-Out Layer Audit}

\begin{table}[H]
\centering
\scriptsize
\renewcommand{\arraystretch}{0.92}
\begin{tabular}{p{0.18\textwidth}p{0.76\textwidth}}
\toprule
Item & Controlled-arity diagnostic setting \\
\midrule
Prompt count & \parbox[t]{0.76\textwidth}{\raggedright 600 prompts per model in the controlled-arity bank, with 100 prompts per arity; the headline diagnostic rows use $r=3,\ldots,6$ (400 prompts per model).} \\
Layer grid & \parbox[t]{0.76\textwidth}{\raggedright 8B: $0,5,10,15,20,25,30,31$; 70B: $20,30,40,50,55,60,70$; 405B: $20,30,40,60,80,90$.} \\
Tested ranks & \parbox[t]{0.76\textwidth}{\raggedright Candidate ranks $k=1,\ldots,7$ span the union of constructor-specific rank grids, but each row is evaluated only on ranks admitted by its active constructor. For argument-only rows, $\mathcal{K}^{\mathrm{arg}}_r=\{1,\ldots,r\}$ and the expected rank is $k=r$. For predicate-plus-argument rows, $\mathcal{K}^{\mathrm{pred+arg}}_r=\{1,\ldots,r+1\}$ and the expected rank is $k=r+1$. Diagonal margins are computed against the other admissible ranks for that constructor, not against nonexistent $k>r$ argument-only tuples.} \\
Tuple budget & \parbox[t]{0.76\textwidth}{\raggedright Headline argument-tuple rows use 20 matched tuples per prompt for true, scrambled, and matched random selectors.} \\
Bootstrap & \parbox[t]{0.76\textwidth}{\raggedright Prompt-level bootstrap confidence intervals use $n=1000$ resamples for the controlled-arity diagnostic summaries.} \\
Held-out audit & \parbox[t]{0.76\textwidth}{\raggedright Even prompt IDs select the displayed layer; odd prompt IDs evaluate the selected row. All 12 headline model/arity rows remain positive on held-out prompts, with held-out confidence intervals excluding zero.} \\
\bottomrule
\end{tabular}
\caption{Compact diagnostic reproducibility summary for the controlled-arity rows. The detailed held-out layer-selection results are shown in Table~\ref{tab:heldout}.}
\label{tab:diagnostic-config}
\end{table}

The layer-selection audit tests whether diagonal rows persist under a held-out layer-selection protocol rather than depending on unrestricted layer search. Table~\ref{tab:diagnostic-config} summarizes the controlled-arity diagnostic configuration used for the headline rows and the held-out audit. Layers are selected on even prompt IDs and evaluated on odd held-out prompt IDs. The held-out set contains 50 prompts per arity row. Table~\ref{tab:heldout} gives the row-level held-out results: every headline model/arity row remains positive with held-out confidence intervals excluding zero.

\begin{table}[!htbp]
\centering
\scriptsize
\renewcommand{\arraystretch}{0.9}
\resizebox{0.86\textwidth}{!}{%
\begin{tabular}{lrrrrrl}
\toprule
Model & $r$ & expected $k$ & dev layer & held-out $D$ & held-out CI & pos. frac. \\
\midrule
8B & 3 & 3 & 25 & 0.645 & [0.612, 0.678] & 1.00 \\
8B & 4 & 4 & 20 & 0.585 & [0.563, 0.604] & 1.00 \\
8B & 5 & 5 & 20 & 0.821 & [0.800, 0.843] & 1.00 \\
8B & 6 & 6 & 20 & 0.959 & [0.938, 0.981] & 1.00 \\
70B & 3 & 3 & 40 & 0.309 & [0.280, 0.334] & 0.98 \\
70B & 4 & 4 & 55 & 0.474 & [0.456, 0.489] & 1.00 \\
70B & 5 & 5 & 40 & 0.653 & [0.619, 0.681] & 1.00 \\
70B & 6 & 6 & 40 & 0.425 & [0.383, 0.467] & 1.00 \\
405B & 3 & 3 & 30 & 0.095 & [0.071, 0.121] & 0.88 \\
405B & 4 & 4 & 30 & 0.226 & [0.200, 0.251] & 1.00 \\
405B & 5 & 5 & 30 & 0.610 & [0.545, 0.668] & 1.00 \\
405B & 6 & 6 & 40 & 0.389 & [0.359, 0.419] & 1.00 \\
\bottomrule
\end{tabular}
}
\caption{Held-out layer-selection validation. Layers are selected on even prompt IDs and evaluated on odd prompt IDs. The held-out $D(r,r)$ remains positive for all headline rows; pos. frac. is the fraction of odd held-out prompts with positive row-level $D(r,r)$.}
\label{tab:heldout}
\end{table}

\section{Prompt Construction}

\subsection{Controlled arity prompts}

Controlled arity prompts contain relation instances with ordered argument spans and distractors. Relation arity $r$ is specified by prompt construction: unary predicates bind one argument, binary relations bind two, and higher-order relations bind larger ordered tuples. For arity $r$, the argument-only tuple constructor enumerates role-order-preserving $k$-subsets for $k<r$ and the full ordered argument tuple for $k=r$; its admissible rank set is $\mathcal{K}^{\mathrm{arg}}_r=\{1,\ldots,r\}$, so it does not create role-only tuples for $k>r$. The predicate-plus-argument constructor prepends the predicate token to the ordered argument tuple; its admissible rank set is $\mathcal{K}^{\mathrm{pred+arg}}_r=\{1,\ldots,r+1\}$, with expected rank $k=r+1$. Scrambled controls use the same constructor and tuple budget after replacing each role token with the same role from another relation in the same prompt. Random controls use matched tuple size and budget but draw unrelated token tuples from the same prompt. Reported diagonal margins compare the expected-rank cell only to the other admissible ranks for that constructor.

\subsection{Edge-grid prompts}

The edge-grid bank uses $8 \times 8$ grids with row tokens $A00,\ldots,A07$ and column tokens $B00,\ldots,B07$. Each row lists the same column field. YES marks an active edge and NO marks an inactive edge. The clean condition uses a coherent target map such as the identity diagonal. The corrupt condition preserves the row/column surface but changes the YES markers to a duplicate-target map. Answer choices include the clean map, the corrupt map, an all-off map, and an all-on map.

A representative clean prompt begins:
\begin{quote}\small
Every A row lists the same B columns in the same order. YES marks the active edge; NO marks inactive edges. Choose the description that fits the YES edges.

Grid: A00: B00=YES B01=NO $\cdots$ B07=NO; A01: B00=NO B01=YES $\cdots$; $\ldots$; A07: $\cdots$ B07=YES.
\end{quote}
The corrupt version preserves the same row/column surface but changes the YES markers to a duplicate-target pattern.

\section{Steering Method Definitions}

Let $C_{\clean}$ and $C_{\corr}$ be clean and corrupt relation-marker clouds at the patch layer, with centroids $\mu_{\clean}$ and $\mu_{\corr}$. Let $\bar C_{\clean}=C_{\clean}-\mu_{\clean}$ and $\bar C_{\corr}=C_{\corr}-\mu_{\corr}$. The centroid is the mean hidden state of the selected marker tokens; the centered cloud $\bar C$ is the relation shape after that mean has been removed. A Procrustes operation fits the orthogonal rotation that best aligns one centered cloud to another. A Grassmann operation instead treats the span of a centered cloud as a subspace and moves that subspace toward a target subspace. Each intervention defines a path $C(\alpha)$ with $\alpha\in[0,1]$ and $C(0)=C_{\corr}$; endpoint definitions below describe $C(1)$.

\begin{itemize}
\item \textbf{linear marker:} add a rowwise clean-minus-corrupt marker delta scaled by $\alpha$.
\item \textbf{centroid+shape:} move the full corrupt relation-marker cloud to the clean relation-marker cloud, so $C(1)=C_{\clean}$.
\item \textbf{shape only:} keep the corrupt centroid but replace the centered cloud shape, so $C(1)=\mu_{\corr}+\bar C_{\clean}$.
\item \textbf{shape\_cross\_prompt\_same\_site:} keep the target corrupt centroid and target marker sites, but use the centered clean shape from a different prompt with the same number of marker rows.
\item \textbf{shape\_cross\_prompt\_corrupt\_same\_site:} keep the target corrupt centroid and target marker sites, but use the centered corrupt shape from a different prompt. This tests whether cross-prompt transfer is clean-state specific.
\item \textbf{shape\_perm\_same\_site:} keep the target corrupt centroid and marker sites, but randomly permute the clean centered marker rows before patching, breaking role/order alignment.
\item \textbf{shape\_reflection\_same\_site:} keep the target corrupt centroid and marker sites, but reflect the clean centered shape through one axis of its local basis before patching.
\item \textbf{clean\_delta\_wrong\_site:} compute the clean-minus-corrupt relation-marker delta, norm-match it, and apply it to matched non-target sites rather than the selected YES/NO marker sites.
\item \textbf{Grassmann shape:} move the subspace spanned by the centered corrupt relation-marker cloud toward the subspace spanned by the centered clean cloud, then reconstruct a marker cloud from that moved shape according to the run configuration.
\item \textbf{centroid+Grassmann shape:} combine the clean-targeted Grassmann shape path with the clean centroid.
\item \textbf{centroid+rotation:} apply the Procrustes-aligned shape and move the centroid to $\mu_{\clean}$.
\item \textbf{Procrustes rotation:} rotate $\bar C_{\corr}$ toward $\bar C_{\clean}$ by the fitted orthogonal alignment while preserving the corrupt centroid.
\item \textbf{spherical marker:} use rowwise spherical interpolation of relation-marker hidden states.
\item \textbf{edge dose:} replace a subset of changed edge markers at each dose.
\item \textbf{Hamming path:} construct an intermediate valid edge map and patch from that intermediate map to the corrupt prompt.
\item \textbf{random rotation:} rotate the centered corrupt cloud toward a random orthogonal target rather than the clean shape.
\item \textbf{random Grassmann:} move toward a random subspace target with matched dimensionality instead of the clean relation subspace.
\item \textbf{random Grassmann matched:} move toward a random Grassmann target while preserving the measured subspace-angle scale.
\item \textbf{equal-norm noise:} add noise with norm matched to the clean-minus-corrupt perturbation scale.
\item \textbf{centroid only:} translate the corrupt relation cloud by $\mu_{\clean}-\mu_{\corr}$.
\item \textbf{random centroid:} translate the corrupt relation cloud by a centroid shift with matched scale but random direction.
\item \textbf{centroid+Grassmann control:} a control variant using a Grassmann-style target without the clean-targeted shape construction used by centroid+Grassmann shape.
\item \textbf{Grassmann rotation only:} rotate toward a Grassmann-derived basis while preserving the corrupt centroid.
\item \textbf{Grassmann basis preserve:} preserve the Grassmann basis structure used by the control while avoiding the clean-targeted shape replacement.
\end{itemize}

The centroid+Grassmann shape path is a clean-targeted shape method; the centroid+Grassmann control row in the full tables is a control variant and is not the same intervention. Control paths use the same patching interface as clean-targeted paths but match only scale, centroid size, rotation, or subspace structure without using the clean relation frame.

\section{Multi-Template Results}

Table~\ref{tab:multitemplate} is included as an audit table for the multi-template summary in Section~3.2. Each row fixes a model, tuple constructor, and arity $r$, then reports the expected-rank gap and strongest off-diagonal margin. Argument-only rows use expected rank $k=r$; predicate-plus-argument rows use expected rank $k=r+1$. A positive margin means that the expected-rank cell is larger than the other tested ranks under the same model, template family, and constructor.

The table shows uniformly positive margins in 405B and constructor-specific mixed cells in 8B/70B. It also records the weaker cells directly, preserving the constructor- and model-specific texture behind the main-text claim.

\begin{table}[H]
\centering
\footnotesize
\setlength{\tabcolsep}{5pt}
\begin{tabular}{llrrrrrr}
\toprule
constructor & $r$ & \multicolumn{2}{c}{8B} & \multicolumn{2}{c}{70B} & \multicolumn{2}{c}{405B} \\
 & & $D$ & margin & $D$ & margin & $D$ & margin \\
\midrule
pred+args & 3 & 0.507 & 0.279 & 0.195 & -0.005 & 0.207 & 0.090 \\
pred+args & 4 & 0.250 & 0.066 & 0.244 & 0.104 & 0.211 & 0.163 \\
pred+args & 5 & 0.331 & 0.215 & 0.309 & 0.190 & 0.188 & 0.049 \\
pred+args & 6 & 0.169 & 0.036 & 0.189 & 0.065 & 0.217 & 0.043 \\
args-only & 3 & 0.549 & 0.167 & 0.403 & 0.138 & 0.320 & 0.195 \\
args-only & 4 & 0.295 & -0.028 & 0.266 & 0.102 & 0.191 & 0.123 \\
args-only & 5 & 0.300 & 0.032 & 0.306 & 0.217 & 0.181 & 0.108 \\
args-only & 6 & 0.444 & 0.216 & 0.150 & 0.001 & 0.240 & 0.169 \\
\bottomrule
\end{tabular}
\caption{Multi-template diagonal margins. Argument-only rows expect $k=r$; predicate-plus-argument rows expect $k=r+1$. In the compact constructor column, args-only denotes argument-only and pred+args denotes predicate-plus-argument.}
\label{tab:multitemplate}
\end{table}

\section{Full Steering Tables}

\begingroup
\scriptsize
\setlength{\tabcolsep}{3pt}
\noindent\textit{70B full steering methods (Table~\ref{tab:70b-full}).}
\vspace{0.35em}
\begin{longtable}{@{}p{0.27\linewidth}ccccc@{}}
\toprule
Method & endpoint beh. & endpoint res. & corr. & coupled AUC & off-target AUC \\
\midrule
\endfirsthead
\multicolumn{6}{l}{Table~\ref{tab:70b-full} continued.}\\
\toprule
Method & endpoint beh. & endpoint res. & corr. & coupled AUC & off-target AUC \\
\midrule
\endhead
linear marker & 1.006 & 0.895 & 0.863 & 0.378 & 0.007 \\
centroid+shape (alias) & 1.006 & 0.895 & 0.863 & 0.378 & 0.007 \\
shape only & 1.006 & 0.895 & 0.860 & 0.375 & 0.007 \\
Grassmann shape & 1.006 & 0.895 & 0.860 & 0.373 & 0.007 \\
centroid+Grassmann shape & 1.006 & 0.895 & 0.863 & 0.376 & 0.007 \\
centroid+rotation & 1.004 & 0.894 & 0.869 & 0.376 & 0.007 \\
Procrustes rotation & 1.006 & 0.895 & 0.866 & 0.372 & 0.007 \\
spherical marker & 1.006 & 0.895 & 0.856 & 0.372 & 0.007 \\
edge dose & 1.006 & 0.895 & 0.775 & 0.211 & 0.006 \\
Hamming path & 1.006 & 0.895 & 0.776 & 0.212 & 0.006 \\
random rotation & 0.478 & 0.742 & 0.861 & 0.073 & 0.009 \\
random Grassmann & 0.130 & 0.612 & 0.918 & 0.010 & 0.009 \\
random Grassmann matched & 0.128 & 0.610 & 0.914 & 0.010 & 0.009 \\
equal-norm noise & 0.024 & 0.016 & 0.146 & 0.000 & 0.008 \\
centroid only & 0.002 & 0.001 & -0.010 & 0.000 & 0.008 \\
random centroid & 0.001 & 0.001 & 0.028 & 0.000 & 0.007 \\
centroid+Grassmann control & 0.005 & 0.003 & 0.027 & 0.000 & 0.007 \\
Grassmann rotation only & 0.008 & 0.002 & 0.055 & 0.000 & 0.008 \\
Grassmann basis preserve & 0.008 & 0.002 & 0.055 & 0.000 & 0.008 \\
\bottomrule
\caption{70B path quality for all steering methods. The main text reports representative families; this table reports every evaluated method with endpoint behavior recovery, endpoint residual blade recovery, behavior-geometry correlation, coupled AUC, and off-target AUC. Linear marker and centroid+shape are alias implementations of the same sampled path in this assay, so identical values are expected.}\label{tab:70b-full}
\end{longtable}
\endgroup

\begingroup
\scriptsize
\setlength{\tabcolsep}{3pt}
\noindent\textit{405B full steering methods (Table~\ref{tab:405b-full}).}
\vspace{0.35em}
\begin{longtable}{@{}p{0.27\linewidth}ccccc@{}}
\toprule
Method & endpoint beh. & endpoint res. & corr. & coupled AUC & off-target AUC \\
\midrule
\endfirsthead
\multicolumn{6}{l}{Table~\ref{tab:405b-full} continued.}\\
\toprule
Method & endpoint beh. & endpoint res. & corr. & coupled AUC & off-target AUC \\
\midrule
\endhead
linear marker & 0.996 & 0.882 & 0.976 & 0.365 & 0.001 \\
centroid+shape (alias) & 0.996 & 0.882 & 0.976 & 0.365 & 0.001 \\
shape only & 0.995 & 0.881 & 0.975 & 0.362 & 0.001 \\
Grassmann shape & 0.995 & 0.881 & 0.974 & 0.361 & 0.001 \\
centroid+Grassmann shape & 0.996 & 0.882 & 0.974 & 0.363 & 0.001 \\
centroid+rotation & 0.996 & 0.877 & 0.977 & 0.359 & 0.001 \\
Procrustes rotation & 0.994 & 0.875 & 0.976 & 0.357 & 0.001 \\
spherical marker & 0.996 & 0.882 & 0.947 & 0.342 & 0.001 \\
edge dose & 0.996 & 0.882 & 0.808 & 0.181 & 0.001 \\
Hamming path & 0.996 & 0.882 & 0.806 & 0.181 & 0.001 \\
random rotation & 0.603 & 0.570 & 0.851 & 0.068 & 0.001 \\
random Grassmann & 0.230 & 0.352 & 0.854 & 0.005 & 0.001 \\
random Grassmann matched & 0.214 & 0.337 & 0.855 & 0.004 & 0.001 \\
equal-norm noise & 0.023 & -0.027 & -0.296 & -0.000 & 0.001 \\
centroid only & 0.003 & -0.000 & -0.011 & 0.000 & 0.001 \\
random centroid & 0.003 & -0.000 & 0.005 & 0.000 & 0.001 \\
centroid+Grassmann control & 0.005 & 0.009 & 0.076 & 0.000 & 0.001 \\
Grassmann rotation only & 0.006 & 0.007 & 0.030 & 0.000 & 0.001 \\
Grassmann basis preserve & 0.006 & 0.007 & 0.030 & 0.000 & 0.001 \\
\bottomrule
\caption{405B path quality for all steering methods. The main text reports representative families; this table reports every evaluated method with the same metrics as Table~\ref{tab:70b-full}. Linear marker and centroid+shape are alias implementations of the same sampled path in this assay, so identical values are expected.}\label{tab:405b-full}
\end{longtable}
\endgroup

\section{Site-and-Ordering Audit Details}

The site-and-ordering audit uses the same edge-grid prompt bank and the same patch/readout interface as the main steering runs: 32 prompts, patch layer 5, readout layer 35, and path fractions $0,0.05,\ldots,1.0$. The 70B run adds order, orientation, and wrong-site placebos to the original shape/centroid/noise controls; the matched 70B and 405B corrupt-donor controls add the clean-state-specific donor placebo under the same patch/readout protocol. Because these rows come from a separately computed control batch, the shape-only rows should be read as within-audit calibration baselines rather than as duplicate copies of the main steering table.

The audit has one especially important contrast. Cross-prompt clean shape succeeds almost as strongly as target clean shape, and in both 70B and 405B the matched cross-prompt corrupt shape fails in coupled recovery. Therefore the donor cloud is not acting as a generic same-site activation source. In the current prompt bank, where the clean map is the identity map for every prompt, the supported interpretation is transfer of a clean/solved identity-frame format across prompt instances and corrupt maps. A stronger claim about transfer across different clean relation contents would require a clean-map-varied prompt bank.

\begingroup
\scriptsize
\setlength{\tabcolsep}{2.8pt}
\renewcommand{\arraystretch}{0.92}
\begin{longtable}{@{}lp{0.34\textwidth}rrrrr@{}}
\toprule
Model & Method & endpoint beh. & endpoint res. & edge rec. & coupled AUC & off-target AUC \\
\midrule
\endfirsthead
\toprule
Model & Method & endpoint beh. & endpoint res. & edge rec. & coupled AUC & off-target AUC \\
\midrule
\endhead
70B & shape only & 1.006 & 0.719 & 1.348 & 0.301 & 0.0067 \\
70B & shape\_cross\_prompt\_same\_site & 1.009 & 0.718 & 1.078 & 0.302 & 0.0068 \\
70B & shape\_cross\_prompt\_corrupt\_same\_site & 0.024 & 0.003 & 0.018 & 0.000 & 0.0055 \\
70B & shape\_perm\_same\_site & 0.140 & 0.309 & 0.551 & 0.023 & 0.0081 \\
70B & shape\_reflection\_same\_site & 0.004 & -0.032 & 0.096 & 0.000 & 0.0074 \\
70B & clean\_delta\_wrong\_site & -0.096 & -0.003 & -0.310 & 0.001 & 0.0072 \\
70B & centroid only & 0.002 & 0.011 & -0.197 & 0.000 & 0.0075 \\
70B & equal-norm noise & 0.024 & 0.006 & 0.256 & 0.000 & 0.0077 \\
\midrule
405B & shape only & 0.991 & 0.852 & 0.992 & 0.358 & 0.0002 \\
405B & shape\_cross\_prompt\_same\_site & 0.992 & 0.847 & 1.009 & 0.365 & 0.0002 \\
405B & shape\_cross\_prompt\_corrupt\_same\_site & -0.001 & -0.002 & 0.078 & 0.000 & 0.0002 \\
405B & shape\_perm\_same\_site & 0.093 & 0.419 & 0.659 & 0.026 & 0.0003 \\
405B & shape\_reflection\_same\_site & 0.005 & -0.004 & 0.170 & -0.000 & 0.0002 \\
405B & clean\_delta\_wrong\_site & -0.040 & 0.016 & 0.434 & 0.000 & 0.0004 \\
405B & centroid only & -0.004 & -0.001 & -0.008 & -0.000 & 0.0002 \\
405B & equal-norm noise & 0.019 & -0.029 & 0.093 & -0.000 & 0.0002 \\
\bottomrule
\caption{Completed site-and-ordering audit. The table reports endpoint behavior, endpoint residual blade recovery, normalized edge-marker Pl\"ucker recovery, coupled AUC, and off-target AUC for 70B and 405B, including matched corrupt-donor controls. Because this is a separately computed site/order control batch, shared method names are audit calibration rows and endpoint residual values should be compared within this audit block rather than row-by-row to Table~\ref{tab:main-steering}. Edge-marker Pl\"ucker recovery can exceed \(1\) when the patched state overshoots the clean reference under this scalar readout.}
\label{tab:site-order-appendix}
\end{longtable}
\endgroup

\section{8B Competence Boundary and Reproducibility Notes}

\subsection{8B competence boundary}

The 8B edge-grid run is retained as a boundary case rather than as primary causal evidence. Baseline clean accuracy is $0.250$, corrupt-answer selection is $0.250$, and the mean clean-minus-corrupt logit gap is $0.029$, so the clean/corrupt answer behavior is not anchored before steering. Some 8B interventions move geometry metrics, but without a reliable behavioral task those movements do not support the main causal claim.

\begin{table}[H]
\centering
\scriptsize
\setlength{\tabcolsep}{4pt}
\begin{tabular}{lccccc}
\toprule
Method & endpoint beh. & endpoint res. & corr. & coupled AUC & off-target AUC \\
\midrule
edge dose & 0.335 & 0.922 & 0.170 & 0.214 & 0.456 \\
linear marker & 0.335 & 0.922 & 0.222 & 0.167 & 0.459 \\
shape only & 0.531 & 0.922 & 0.217 & 0.155 & 0.459 \\
random rotation & -0.297 & 0.818 & -0.124 & -0.081 & 0.454 \\
centroid only & 0.041 & -0.000 & 0.040 & 0.000 & 0.456 \\
equal-norm noise & -0.339 & 0.064 & -0.044 & -0.008 & 0.456 \\
\bottomrule
\end{tabular}
\caption{8B edge-grid steering metrics are not used as main causal evidence because the baseline task is not solved. Off-target AUC is high, indicating poor answer localization.}
\label{tab:8b-boundary}
\end{table}

\subsection{Core run configuration}

Main steering runs used compact edge-grid prompts, corrupt mode ``broken,'' 32 prompts, 8 nodes, patch layer 5, path fractions $0,0.05,\ldots,1.0$, projection dimension 64, projection seed 42, tuple budget 8, subspace dimension 8, and 300 bootstrap resamples. The site-and-ordering audits used the same patch/readout setting with 500 bootstrap resamples where intervals are computed. The 70B and 405B main readout layer was 35. The 8B boundary run used readout layer 20. The main 405B run was merged from four chunks and passed basic quality checks: 12,768 rows, 32 prompts, 19 method rows including the linear-marker/centroid+shape alias, no duplicate prompt-method-path-fraction rows, and no missing behavior, residual, or answer-probability fields. The matched 70B corrupt-donor control contains 672 rows over 32 prompts and 21 path fractions. The final 405B site-and-ordering controls were merged from two chunks with 5,376 rows over 32 prompts and passed the same duplicate-row and missing-field checks.

\begin{table}[!htbp]
\centering
\small
\begin{tabular}{@{}p{0.25\linewidth}p{0.67\linewidth}@{}}
\toprule
Setting & Value \\
\midrule
Prompt bank & compact edge-grid, corrupt mode \texttt{broken} \\
Prompts & 32 \\
Nodes & 8 A-rows by 8 B-columns \\
Patch site & selected YES/NO marker token positions \\
Patch layer & 5 \\
Readout layer & 35 for 70B/405B \\
Path fractions & $0,0.05,\ldots,1.0$ \\
Methods & 19 evaluated method rows, including one alias path; site-order audits for 70B/405B \\
Projection & deterministic Gaussian, dimension 64, seed 42 \\
Subspace dimension & 8 \\
Tuple budget & 8 \\
Large-model loading & 4-bit BitsAndBytes with bfloat16 compute for 70B/405B \\
Bootstrap & 300 prompt-level resamples for main steering; 500 for site-order audits \\
\bottomrule
\end{tabular}
\caption{Core steering configuration for the reported 70B and 405B experiments.}
\label{tab:repro-config}
\end{table}
\FloatBarrier

\end{document}